\documentclass[twoside,11pt]{article}

\usepackage{jmlr2e}

\usepackage{amsmath}
\usepackage{amsfonts}
\usepackage{xr}
\usepackage{accents}

\newcommand{\argmin}{\operatornamewithlimits{argmin}}

\newcommand{\half}{\frac{1}{2}}
\newcommand{\RR}{\mathbb{R}}
\newcommand{\NN}{\mathbb{N}}

\newtheorem{assumption}{Assumption}

\usepackage{caption} 

\firstpageno{1}

\begin{document}

\title{Shift-Curvature, SGD, and Generalization}

\author{\name Arwen V. Bradley \email arwen\_bradley@apple.com 
    \AND
    \name Carlos A. Gomez-Uribe \email cgomezuribe@apple.com 
    \AND
    \name Manish Reddy Vuyyuru \email mvuyyuru@apple.com \\
    \addr Apple, Inc.  1 Infinite Loop, Cupertino, CA, USA.}


\maketitle

\begin{abstract}
A longstanding debate surrounds the related hypotheses that low-curvature minima generalize better, and that SGD discourages curvature. We offer a more complete and nuanced view in support of both. First, we show that curvature harms test performance through two new mechanisms, the shift-curvature and bias-curvature, in addition to a known parameter-covariance mechanism. The three curvature-mediated contributions to test performance are reparametrization-invariant although curvature is not. The shift in the shift-curvature is the line connecting train and test local minima, which differ due to dataset sampling or distribution shift. Although the shift is unknown at training time, the shift-curvature can still be mitigated by minimizing overall curvature. Second, we derive a new, explicit SGD steady-state distribution showing that SGD optimizes an effective potential related to but different from train loss, and that SGD noise mediates a trade-off between deep versus low-curvature regions of this effective potential. Third, combining our test performance analysis with the SGD steady state shows that for small SGD noise, the shift-curvature may be the most significant of the three mechanisms. Our experiments confirm the impact of shift-curvature on test loss, and further explore the relationship between SGD noise and curvature.
\end{abstract}


\section{Introduction}

In typical machine learning applications, we train a model on a training dataset hoping that the model will perform well on a test dataset, which is a proxy for unseen data. However, our theoretical understanding of how datasets, models, and learning algorithms combine to determine test performance is still incomplete, particularly in the case of overparametrized models that defy classical expectations about overfitting (\cite{belkin2021fit}). 
A decades-old debate focuses on the related hypotheses that low curvature of the loss function results in better test performance, and that stochastic gradient descent (SGD) favors lower-curvature local minima. There is strong evidence for these hypotheses. First, many studies have shown that increasing SGD noise (by increasing the ratio of learning rate to batch size) can improve test performance (\cite{goyal2017accurate, he2019control, smith2020generalization, mccandlish2018empirical, golmant2018computational, hoffer2017trainlonger, shallue2018measuring, you2017scaling}). Our experiment in Figure \ref{fig:vgg_lb_init_hpuh5hk7ja} also confirms this. Second, \cite{keskar2016large, jiang2019fantastic, li2017visualizing} offer empirical evidence that lower-curvature local minima tend to generalize better, an idea often attributed to \cite{hochreiter1997flat}. Third, \cite{jastrzkebski2017three} make direct theoretical connections between test performance, curvature, and SGD noise (albeit under strong assumptions). They show that under the SGD steady-state distribution, the expected test loss near a given local minimum depends on the trace of curvature times parameter covariance, assuming a constant Hessian equal to SGD noise. They also show that the probability of SGD landing near a particular local minimum is inversely related to its curvature, with noise mediating a depth/curvature tradeoff, assuming constant, isotropic SGD noise and valid Laplace approximation. On the other hand, there are several arguments against the idea that curvature predicts test performance. \cite{dinh2017sharp} point out that model reparametrization can arbitrarily change curvature without changing test performance. Further, a number of recent works argue that \cite{jastrzkebski2017three} and others present an oversimplified view of curvature, since loss landscapes often (e.g., when there are many more model parameters than training datapoints) have many directions of near-zero curvature rather than locally strictly-convex valleys; in particular, the SGD noise matrix and Hessian of the loss are different, and both are highly non-isotropic with many near-zero eigenvalues (\cite{sagun2018empirical, li2020hessian, draxler2018essentially}). Our work aims to help reconcile these points of view by removing unrealistic assumptions such as constant, isotropic Hessian and loss gradient covariance, identifying new and possibly more significant mechanisms by which curvature affects test performance, and showing that these curvature-dependent contributions to test performance are reparametrization invariant.

\begin{figure}[ht]
\centering
\small{VGG16}
\includegraphics[width=1.0\columnwidth]{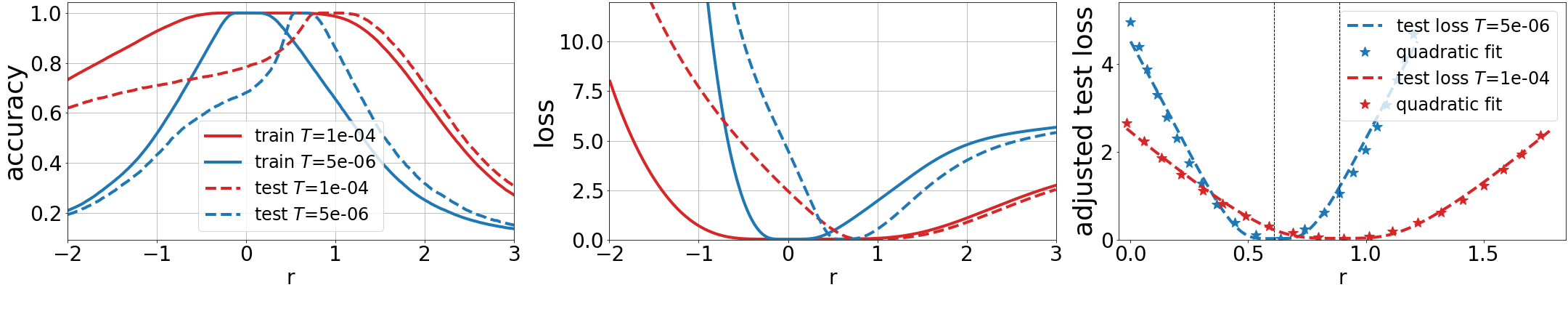} \\
\small{Resnet10}
\includegraphics[width=1.0\columnwidth]{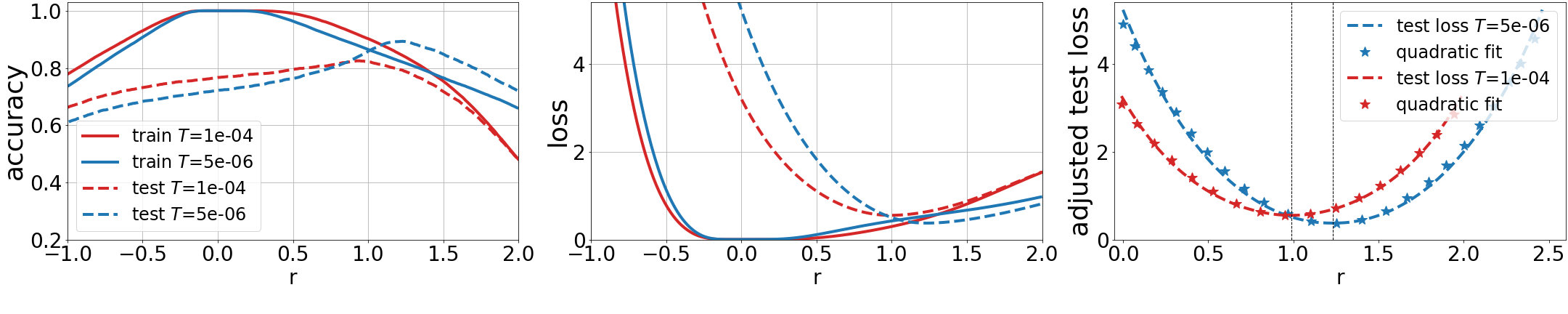}
\caption{Train and test accuracy (left) and loss (middle) along a line connecting the local minimum of the train loss to a nearby local minimum of the test loss, for different train local minima found using high (red) vs. low (blue) temperature SGD. Lower temperature SGD leads to higher curvature (for both train and test, in terms of both loss and accuracy), and worse test performance near the train local minimum (compare red to blue dashed lines near $r=0$). We extract loss curvature by reflecting the train and test losses along the line about their respective local minima and fitting the resulting curve to $\half c x^2$. We call $c$ the curvature; the right-most plot above shows the reflected losses and the quadratic fits corresponding for a specific temperature. In all plots, the $x$- and $y$-axes are as follows: letting $\Theta(r)$ be a line connecting the train/test local minima (so that $\Theta(0) = \theta^\textit{tr}_k$ and $\Theta(||s_k||) = \theta^\textit{test}_k$), the $x$-axis is $r$, and $y$-axis is the accuracy or loss evaluated at the interpolated parameters $\Theta(r)$.
}
\label{fig:train_test_loss}
\end{figure}

At a high level, we find that the intuitive connections between test performance, curvature, and SGD noise still hold, but in a more complete and nuanced way. We show that the test loss averaged over an arbitrary parameter distribution depends on the \emph{shift-curvature} (in addition to the \emph{bias-curvature} and the previously-identified covariance-curvature). The shift-curvature is equal to the curvature of the loss in the direction of the \emph{shift} between nearby train and test local minima, as shown in Figure \ref{fig:train_test_loss}; such a shift may be caused by dataset sampling or possibly by test and train datasets coming from different distributions. However, the shift is unknown at training time, so it is beneficial to reduce curvature in all directions. Second, we show that the SGD steady-state distribution favors train local minima with low curvature in all directions, although the curvature is not that of the train loss, but rather of a related \emph{effective potential} which we obtain by refining existing results about the steady state of SGD. Third, we find that the shift-curvature may be the most significant curvature term when SGD noise is small. Our empirical results show that shift-curvature indeed has a significant impact on test loss.

In more detail, we refine existing results for the steady-state distribution of SGD as follows. Starting from a continuous-time diffusion approximation of SGD due to \cite{li2017stochastic} (for which we also offer an alternative, constructive derivation which may be helpful for analyzing other algorithms), we derive a steady-state distribution showing that SGD minimizes an explicit effective potential that is related to but different from the training loss. Our expression is more explicit than the result in \cite{chaudhari2018stochastic}, and more general than that in \cite{jastrzkebski2017three} and \cite{seung1992statistical} since we remove their assumptions of constant and isotropic SGD noise. 

This paper is organized as follows. Sections \ref{sec:setup} and \ref{sec:main_results} present the setup and main results for the simpler underparametrized case (more data samples than model parameters). Section \ref{sec:sgd} discusses the SGD diffusion approximation and its steady state in more detail, describing how the results in Section \ref{sec:main_results} are obtained. Section \ref{sec:overparam} extends the results to the more complex overparametrized case. The appendices contain derivations and additional experiments.

\begin{figure}[t]
\centering
\includegraphics[width=0.49\columnwidth]{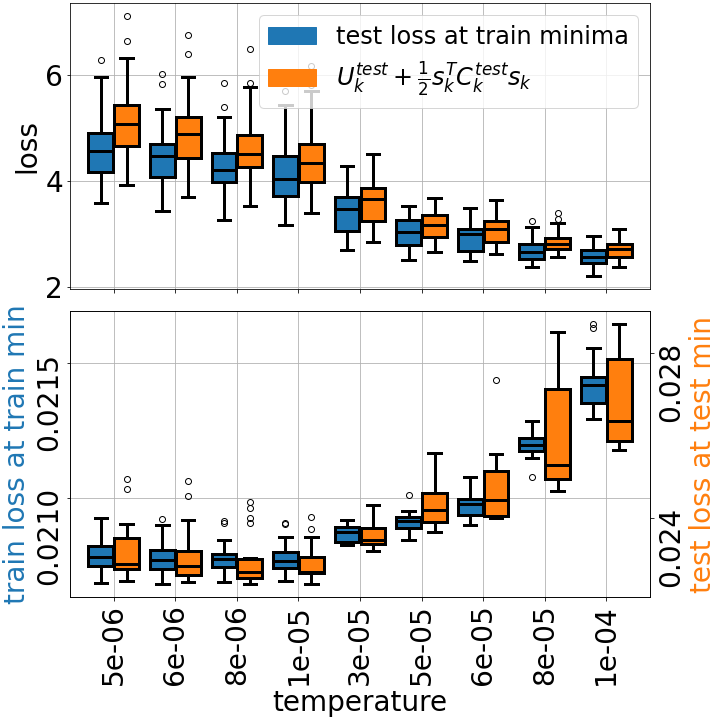}
\includegraphics[width=0.49\columnwidth]{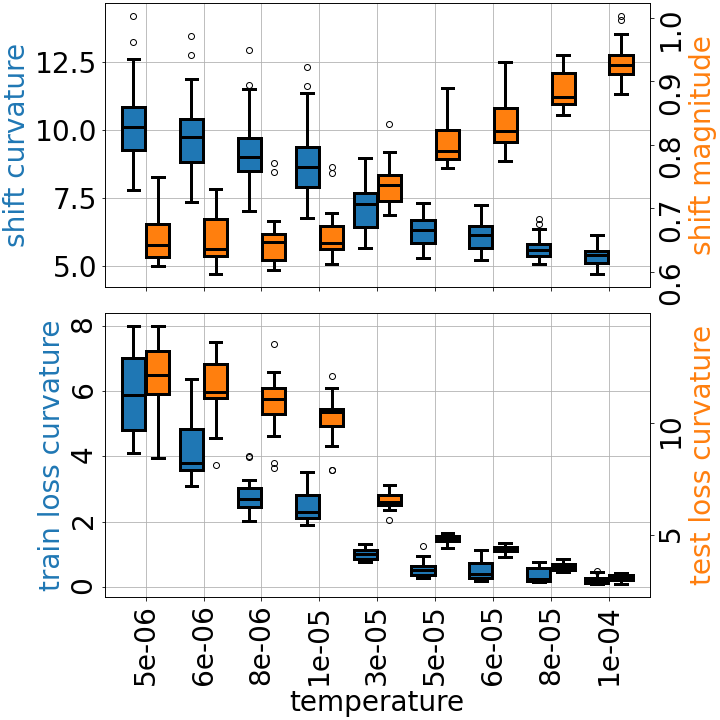}
\caption{Test loss and shift-curvature for VGG16 network. Experiment setup as described in Figure \ref{fig:train_test_loss}. (Top left) Test loss at the train local minimum and Taylor approximation prediction as a function of temperature. The former is an important quantity because it predicts model performance on unseen data; note the improvement with increasing temperature. Equation \ref{eq:taylor_test_loss} (with $\hat \theta = \theta^\textit{tr}_k$) predicts that the test-loss-at-train-local-minimum is approximately equal to the test-loss-at-test-local-minimum $U^{\textit{test}}_k$ plus the shift-curvature term $\half s_k C^\textit{test}_k s_k$, which is supported by our experimental results. (Bottom left) Train loss and test loss evaluated at their respective local minima ($U^{\textit{tr}}_k$, $U^{\textit{test}}_k$), as a function of temperature. Both increase with increasing temperature, consistent with the theory that SGD temperature trades off depth and curvature. (Top right) Shift-curvature ($s_k C^\textit{test}_k s_k$) and shift magnitude ($\|s_k\|$). Shift-curvature decreases with increasing temperature, as expected. We observe empirically that $s_k C_k^\textit{test} s_k$ makes a larger contribution to (\ref{eq:taylor_test_loss}) than $U^{\textit{test}}_k$. Finally, here the shift magnitude increases with temperature, though we see the opposite behavior for Resnet10 (Figure \ref{fig:resnet_nbn_shift_curv}). (Bottom right) Train and test curvature ($C_k^\textit{train}, C_k^\textit{test}$), which both decrease with increasing temperature as expected.}
\label{fig:vgg_shift_curv}
\end{figure}

\begin{figure}[t]
\centering
\includegraphics[width=0.49\columnwidth]{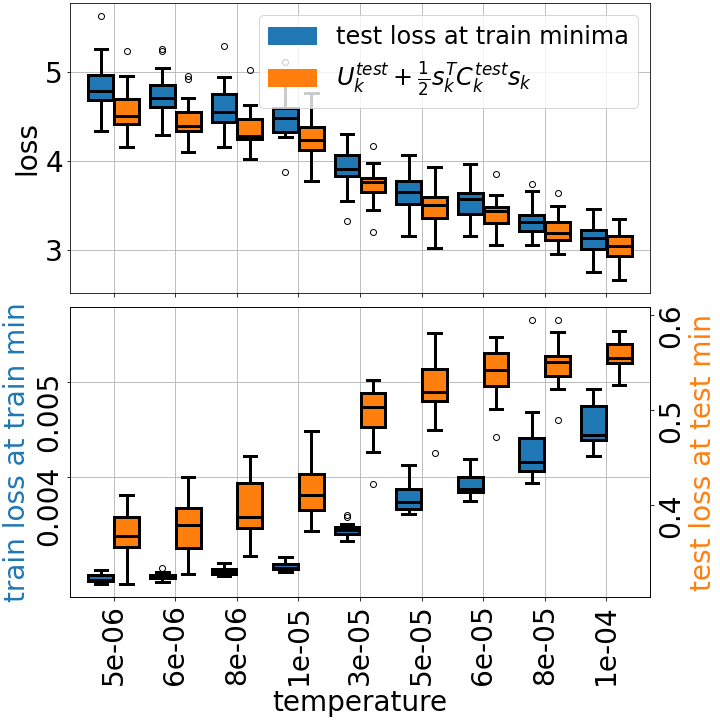}
\includegraphics[width=0.49\columnwidth]{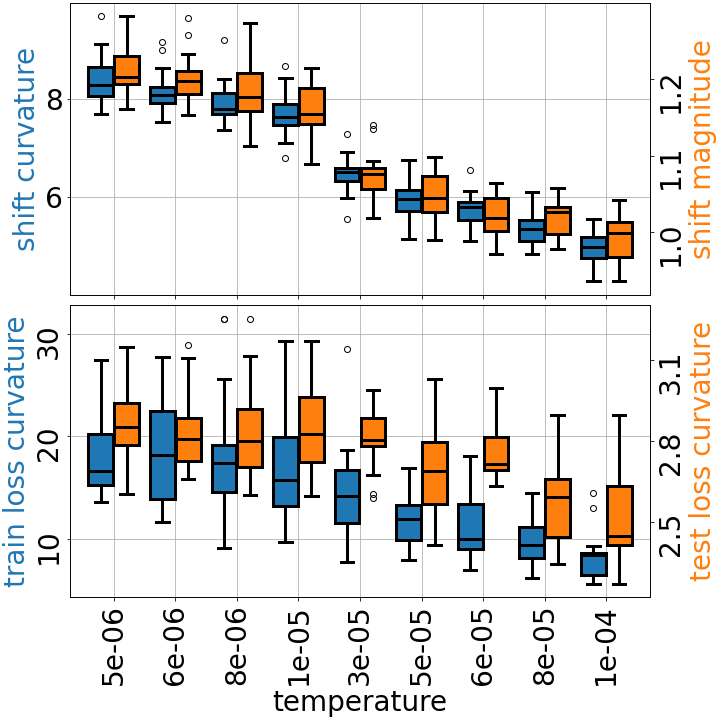}
\caption{Experiment as in Figure \ref{fig:vgg_shift_curv}, repeated for Resnet10 network. The main qualitative differences are that the shift now decreases with increasing temperature, and the magnitude of the train curvature is higher than that of the test curvature.}
\label{fig:resnet_nbn_shift_curv}
\end{figure}

\section{Setup} \label{sec:setup}
In this section and the next, we present our setup and main results in the underparametrized case (Section \ref{sec:overparam} extends to the overparametrized case). We assume we are given a \textit{loss} function $U(x, \theta)$ that depends on a single data sample $x \in \RR^\iota$ and model parameters $\theta \in \RR^p$, where $\iota$ and $p$ are positive integers. We are also given a training set $\{x^{\textit{tr}}_1, \ldots, x^{\textit{tr}}_{N_\textit{tr}} \}$ (\textit{tr} is short for train) consisting of $N_\textit{tr}$ points sampled i.i.d. from an underlying train distribution, and a test set $\{x^{\textit{test}}_1, \ldots, x^{\textit{test}}_{N_{\textit{test}}} \}$ sampled i.i.d. from an underlying test distribution. Often the train and test distributions are identical, but sometimes they differ from one another (for example, when the data distribution evolves over time and test is sampled at a later time than train). But even when train and test are both sampled from the same underlying distribution, the train and test \textit{sets} still differ due to sampling (because each data set consists of a finite number of samples). We define the train and test losses for a given $\theta$ as the expectation over the train and test sets, respectively:
\begin{align}
    U^{\textit{tr}}(\theta) = \frac{1}{N_\textit{tr}}\sum_{i=1}^{N_\textit{tr}}U(x^{\textit{tr}}_i, \theta),  \text{ and }
    U^\textit{test}(\theta) = \frac{1}{N_\textit{test}}\sum_{i=1}^{N_\textit{test}}U(x^{\textit{test}}_i, \theta), \label{eq:losses}
\end{align}
In general, $U^\textit{tr}(\theta)$ and $U^{\textit{test}}(\theta)$ are similar but not identical functions, each with multiple local minima. We make three initial assumptions about the local minima:
(i) the local minima of both train and test are strict,
(ii) each local minimum of the train loss has a nearby local minimum of the test loss, and
(iii) the train local minima are countable.
We expect (i) to hold for underparametrized models (provided there are at least $p$ linearly-independent per-sample gradients) but not for overparametrized models (which are handled in Section \ref{sec:overparam}, where we drop assumption (i)). Assumption (ii) is validated by our experiments, and can be satisfied, for example, in the absence of distribution shift by having large enough datasets. Both (ii) and (iii) could actually be relaxed to apply only to train local minima that receive significant weight in the SGD steady-state distribution, as (\ref{eq:approxAvgTestLoss}) will make clear, but we will assume they hold for all train minima to simplify the exposition. With these assumptions, we denote the (strict) local minima of train by $\theta^\textit{tr}_k$ and the corresponding closest local minima of test by $\theta^{\textit{test}}_k$, and generally reserve $k$ as an index that runs through train local minima ($k = 1, 2, 3, \ldots$). We can then define the \emph{shift} between corresponding local minima of train and test as:
\begin{equation}
    s_k=\theta^{\textit{test}}_k - \theta^\textit{tr}_k
    \label{eq:shift}
\end{equation}
The shift is small as a consequence of assumption (ii), but nonzero in general because the train and test losses are slightly different (due to dataset sampling or possibly distribution shift). \cite{he2019asymmetric}'s experiments and our own demonstrate a nonzero train/test shift for several models and datasets. (We generalize definition (\ref{eq:shift}) to apply to non-strict minima in Section \ref{sec:overparam}.) Table \ref{table:symbols} summarizes the main notation we use for easy reference.

We further assume that a stochastic learning algorithm such as SGD processes the train dataset to produce model parameters with a distribution $\rho(\theta);$ that is, different runs of SGD on the same training set yield i.i.d. samples from $\rho(\theta).$ The resulting models are evaluated on a test dataset to determine the test performance.
One of our main goals is to understand the expected test loss over the distribution of model parameters the algorithm produces (while keeping datasets fixed). 

\begin{table}[t]
\centering
\begin{tabular}{|c c c|} 
 \hline
 Symbol & Definition & Notes \\ [0.5ex] 
 \hline
 $U(x, \theta)$ & & Loss function\\
 $U^{\textit{tr/test}}(\theta)=$ & $E_{x \sim \textit{tr/test}}[U(x, \theta]$ & Train/test loss at $\theta$ \\ 
 $\theta^{\textit{tr/test}}_k=$ & $\argmin U^{\textit{tr}/\textit{test}}(\theta)$ & Local minima of test/train loss \\
 $U^{\textit{tr/test}}_k$ & $U^{\textit{tr}/\textit{test}}(\theta_k^{\textit{tr}/\textit{test}})$ & Test/train loss at local minima \\ 
 $C^{\textit{tr/test}}_k$ & $\partial^2_\theta U^{\textit{tr}/\textit{test}}(\theta_k^{\textit{tr}/\textit{test}})$ & Curvature at test/train local minima  \\
 $s_k=$ & $\theta^{\textit{test}}_k - \theta^\textit{tr}_k$ & Shift between train/test minima \\
 $\rho(\theta) \approx$ & $\sum_k w_k \mathcal{N}\big(\mu_k, \Sigma^2_k \big)$ & Approx. parameter distribution \\
 $b_k=$ & $\theta^\textit{tr}_k - \mu_k$ & Bias of $\rho$ at $k$ \\
 $D^\textit{tr}(\theta) =$ & $\text{Cov}_{x \sim \textit{tr}}[\partial_\theta U(x, \theta)]$ & SGD diffusion matrix \\
 [1ex]
 \hline
\end{tabular}
\caption{Notation reference guide}
\label{table:symbols}
\end{table}

\section{Main Results} \label{sec:main_results}
First, consider a solution $\hat{\theta}$ obtained by a single run of a training algorithm like SGD, i.e., $\hat{\theta}$ is a sample from $\rho(\theta)$. To predict the performance of the trained model on unseen data, we would evaluate the test loss at $\hat \theta$. We expect $\hat{\theta}$  to be near a train local minimum $\theta^\textit{tr}_k$ because it was found by an algorithm that attempts to minimize the train loss. We can therefore write 
$\hat \theta = \theta^{\textit{test}}_k - s_k - \hat b,$
where $s_k$ is the train/test shift introduced above, and $\hat{b} = \theta^\textit{tr}_k - \hat{\theta}$ denotes the \emph{bias} of $\hat{\theta}$ relative to the train local minimum, with both $s_k$ and $\hat{b}$ assumed small. We then Taylor expand $U^{\textit{test}}$ about $\theta^{\textit{test}}_k$, noting that $\partial_\theta U^{\textit{test}}(\theta^{\textit{test}}_k) = 0$:
\begin{align}
    U^{\textit{test}}(\hat \theta) &\approx
    U^{\textit{test}}_k + \half (s_k + \hat b)^T C^{\textit{test}}_k (s_k + \hat b), \label{eq:taylor_test_loss}
\end{align} 
where we introduce the more compact notation $U^{\textit{test}}_k = U^{\textit{test}}(\theta^{\textit{test}}_k)$ for the test loss at a local minimum, and $C^{\textit{test}}_k = \partial^2_\theta U^{\textit{test}}(\theta^{\textit{test}}_k)$ for the Hessian (or \emph{curvature}) of the test loss at one its local minima.\footnote{In Section \ref{sec:overparam}, we find that (\ref{eq:taylor_test_loss}) still holds in the overparametrized case after generalizing the definition of $s_k$ to apply to non-strict minima. We further note that for asymmetric losses, only the curvature in the direction from the test toward the train local minimum matters, as explored in our experiments. Finally, in Appendix \ref{append:higher_order}, we show how this result and others can be partially generalized to the case where the quadratic term is zero, but curvature can still be characterized in terms of a higher-order even derivative.} Although it comes from a simple Taylor-expansion, Equation \ref{eq:taylor_test_loss} identifies important quantities impacting test loss that have so far been inadequately discussed in the literature. The equation says that the test loss evaluated near a train local minimum is equal to the \emph{test loss at the test local minimum} plus a term that depends on an interaction between the \emph{curvature}, the \emph{shift} between the train and test local minima, and the \emph{bias} between $\hat{\theta}$ and the train local minimum. The \emph{shift-curvature} term, $\half s_k C^{\textit{test}}_k s_k$, plays an especially important role in the rest of this paper; the basic intuition is that, given train and test losses with local minima slightly shifted relative to each other, the test loss evaluated near a train local minimum depends on the local curvature in the direction of the local-minimum shift (\cite{keskar2016large} seem to appeal to similar intuition, but do not make it concrete). Since the shift cannot be known at training time, it is beneficial to minimize curvature in all directions, which, as we will see shortly, is what SGD does.

In order to obtain an explicit approximation for the expected test loss over a distribution of parameters, we model the parameter distribution $\rho(\theta)$ as a countable mixture of Gaussians with one mixture component for each train loss minimum, as proposed by \cite{jastrzkebski2017three} for the case of SGD. That is, we write $\rho(\theta)=\sum_k w_k \mathcal{N}\big(\mu_k, \Sigma_k \big)$, where $w_k \geq 0$, $\sum_k w_k=1$, the component means $\mu_k$ are close to the train local minima, $b_k=\theta^\textit{tr}_k - \mu_k$ denote the component \emph{biases}, and $\Sigma_k$ denote the component covariances.\footnote{Generally, such a model is reasonable when $\rho$ has the form $\rho(\theta) \propto e^{-cv(\theta)}$ for some constant $c$ and function $v$ with local minima near those of the train loss, so that $\rho$ is multimodal with peaks near the train local minima, enabling a Laplace approximation about each local minimum of $v$, which yields a Gaussian mixture with small biases. We will see shortly that the SGD steady state has this form.} Averaging the approximate test loss in (\ref{eq:taylor_test_loss}) over such a parameter distribution results in the approximate test performance
\begin{align}
    E_{\theta \sim \rho}[U^{\textit{test}}(\theta)] \approx & \sum_k w_k \big\{U^{\textit{test}}_k + \half \text{Tr} [\Sigma_k C^{\textit{test}}_k] + 
    \half (s_k + b_k)^T C^{\textit{test}}_k (s_k + b_k)\big\}.\label{eq:approxAvgTestLoss}
\end{align} 
Here $\text{Tr}[\cdot]$ denotes the trace of a matrix. Equation \ref{eq:approxAvgTestLoss} is one of our main results, and makes explicit how curvature determines test performance: through the covariance of model parameters $\Sigma_k,$ as well as a quadratic function of the shifts $s_k$ and biases $b_k$. \cite{jastrzkebski2017three, buntine1991bayesian} identify the same $\text{Tr} [\Sigma_k C^{\textit{test}}_k]$ contribution of curvature to test performance (under the assumption that $C = D$, which we remove), but the other curvature-dependent terms in (\ref{eq:approxAvgTestLoss}) are new as far as we know. In particular, the new shift-curvature term (discussed above) appears as $\half w_k s_k C^{\textit{test}}_k s_k$, showing that test loss improves when $\rho$ places more weight on minima with lower shift-curvature.

We can also show that (\ref{eq:approxAvgTestLoss}) is reparametrization-invariant. Although multiple authors, including \cite{jiang2019fantastic}, \cite{keskar2016large}, and \cite{hochreiter1997flat}, have suggested that generalization quality can be tied directly to curvature, \cite{dinh2017sharp} argue against this by observing that model reparametrization can arbitrarily change the curvature of the local minima. Although the curvature alone is indeed not invariant to reparametrization, we show in Appendix \ref{append:reparam} that every term in (\ref{eq:approxAvgTestLoss}) \emph{is} reparametrization-invariant, as is the entire expression. The intuition behind the proof is that reparametrization changes terms in (\ref{eq:approxAvgTestLoss}) like $s_k$ and $C^{\textit{test}}_k$ in ways that cancel each other out, i.e., under a reparametrization $y=r^{-1}(\theta)$, the reparametrized shift becomes $\partial_y r(y_k^\textit{test})^{-1} s_k$ while the reparametrized curvature becomes $\partial_y r(y_k^\textit{test})^T C^{\textit{test}}_k \partial_y r(y_k^\textit{test})$ (where $\partial_y r(y)^{-1}$ is the matrix inverse of the Jacobian of $r$), so that $s_k C^{\textit{test}}_k s_k$ is left unchanged. Other authors like \cite{smith2017bayesian} have shown reparametrization-invariant PAC-Bayes bounds for the generalization error, but we show the left-hand-side of (\ref{eq:approxAvgTestLoss}) is a reparametrization-invariant quantity that directly approximates the expected test loss.

Our second main result is the approximate steady-state distribution of SGD, derived in Section \ref{sec:sgd}:
\begin{align}
    \rho(\theta) &=\frac{1}{Z} \text{exp} \biggr\{- \frac{2}{T} v(\theta) \biggr\}, \quad \text{where } Z \equiv \int \text{exp} \biggr\{- \frac{2}{T} v(\theta) \biggr\} d\theta, \text{ and} \nonumber\\
    v(\theta) &\equiv \int^{\theta} (D^\textit{tr})^{-1} \big( \partial_\theta U^\textit{tr}  +  \frac{T}{2}(\partial_{\theta} \cdot D^\textit{tr} )^T \big) \cdot d \theta. \label{eq:sgdss}
\end{align}
We call $v(\theta)$ the \textit{effective potential}, $D^\textit{tr}(\theta)$ is the training set gradient covariance, $\partial_\theta U^\textit{tr}(\theta)$ is the gradient of the training loss, and $Z$ is the partition function of this Boltzmann-like distribution. The \emph{temperature} $T$, which is equal to the ratio of learning rate and minibatch size (so typically $T \ll 1$), captures the strength of SGD noise (as discussed further in Section \ref{sec:sgd}). The effective potential is related to the training loss $U^\textit{tr}(\theta)$, but in general not the same: $v \propto U^\textit{tr}$ only if $D^\textit{tr}$ is constant and isotropic.\footnote{When $D^\textit{tr} = d(\theta) I$ for some scalar function $d(\theta)$, then $ v(\theta) =  \int^\theta \frac{1}{d(\theta)}\partial_\theta U^\textit{tr} \cdot d\theta + \frac{T}{2}\log(d(\theta))$. If in addition $d(\theta)$ is a constant denoted by $d,$ then $v(\theta) = d^{-1} U^\textit{tr}(\theta) + \text{const}.$}
Result \ref{eq:sgdss} requires the assumption that the integrand in the line integral is a gradient, which is equivalent to having zero curl, as well as for $D^\textit{tr}(\theta)$ to be invertible, which is only true in underparametrized case.\footnote{The zero-curl assumption is stated and discussed in Appendix \ref{append:steady-state} (Assumption \ref{assump:curl}). The invertibility of $D^\textit{tr}$ is handled in Section \ref{sec:overparam}, where we find that an analog of (\ref{eq:sgdss}) still holds in the overparametrized case if we add zero-mean Gaussian noise to SGD and $\ell_2$ regularization to the training loss.} Equation \ref{eq:sgdss} is more general than the results in \cite{jastrzkebski2017three, seung1992statistical}, which assume that the noise $D^\textit{tr}(\theta)$ is constant and isotropic (i.e., a multiple of the identity). It is different and more explicit than results in \cite{chaudhari2018stochastic} for nonconstant, anisotropic noise, enabled by the zero-curl assumption that we rely on to obtain a solution with zero probability current ($J$ in Eq. (\ref{eq:fp})). \cite{chaudhari2018stochastic} say that the steady state cannot have zero probability current  for anisotropic and/or non-constant noise, but we offer a counterexample in Appendix \ref{append:steady-state}. They also stop short of explicitly solving for the effective potential, presumably because they are rightly concerned about inverting $D^\textit{tr}$; however, our analysis of the nullspace of $D^\textit{tr}$ shows that this is only an issue in the overparametrized case, in which case a modified SGD still allows us to find a solution (see Section \ref{sec:overparam}).

Approximating $\rho$ as a mixture of Gaussians shows that $\rho$ assigns to basin $k$ the weight  
\begin{align}
   w_k &\propto  \text{exp} \{ -\frac{2}{T} v(\mu_k,T) \} \cdot | \partial_\theta^2 v(\mu_k, T) |^{-\half}. \label{eq:weight_rho}
\end{align}
This shows how SGD noise mediates a trade-off between depth and curvature of the effective potential, generalizing the result given in \cite{jastrzkebski2017three} for the special case of constant, isotropic covariance $D^\textit{tr}(\theta)$.
When $T = 0$ the weights $w_k$ indicate that $\rho(\theta)$ places all its probability mass on the deepest minimum of the effective potential, but as $T$ increases, $\rho(\theta)$ begins to transfer weight onto other minima. It then favors minima with lower curvature of the effective potential due to the inverse-square-root dependence on the determinant of the curvature of the effective potential. In particular, if two local minima have the same depth, SGD will the prefer the one with lower curvature. Compared to \cite{jastrzkebski2017three}'s result, ours highlights the complexity arising from non-isotropic noise, in particular that the depth and curvature are those of the \emph{effective potential} rather than those of the training loss.

Third, to approximate the expected SGD test performance, we specialize (\ref{eq:approxAvgTestLoss}) to the Gaussian mixture approximation of the SGD steady-state $\rho$ in (\ref{eq:sgdss}) to obtain
\begin{align}
    E_{\theta \sim \rho}[U^{\textit{test}}(\theta)] &\approx \sum_k w_k (U^{\textit{test}}_k + \half s_k^T C^{\textit{test}}_k s_k) + O(T). \label{eq:testPerfSGDApprox}
\end{align}
The terms in (\ref{eq:approxAvgTestLoss}) involving bias $b_k$ and covariance $\Sigma_k$ are both $O(T)$, as detailed in Section \ref{sec:sgd}. Equation \ref{eq:testPerfSGDApprox} shows that for SGD with small $T$ and non-negligible shifts $s_k$, our newly-identified shift-curvature term may be the most significant way that curvature affects the test loss. In summary, we find that SGD temperature mediates a trade-off between depth and curvature of the effective potential, so that as temperature increases, greater weight $w_k$ is placed on lower-curvature basins. These lower-curvature basins tend to have smaller values of shift-curvature, $s_k^T C^{\textit{test}}_k s_k$, which can reduce the expected test loss $E_{\theta \sim \rho}[U^{\textit{test}}(\theta)]$ overall. We illustrate this with a conceptual example in the next section.

Our experiments in Figures \ref{fig:train_test_loss}, \ref{fig:vgg_shift_curv}, \ref{fig:resnet_nbn_shift_curv} demonstrate the significant contribution of shift-curvature to test performance, support the inverse dependence of curvature on SGD temperature, and offer additional empirical insights about the test local minima, curvature, and shift. For Figures \ref{fig:train_test_loss}, \ref{fig:vgg_shift_curv}, \ref{fig:resnet_nbn_shift_curv}, we trained VGG16 (\cite{simonyan2014very}) and Resnet10 (\cite{he2016deep}) networks, both without batch normalization, on CIFAR10.
We repeated the following experiment over a range of temperatures for the initial SGD run (and over several seeds for each temperature). For each temperature and seed, we first trained our model using SGD with the specified temperature to get close to local minima of train (note that SGD typically finds \emph{different} train local minima for different temperatures). Next, we initialized from the SGD solution and kept training at extremely low temperature on the train set to get even closer to the train local minimum. Finally, we again initialized from the SGD solution and trained at extremely low temperature, but this time on the \emph{test set}, to get as close as possible to the test local minimum. Having found the train and test local minima, we linearly interpolated between them in order to study the train and test accuracy and loss along the line connecting their local minima (as shown in Figure \ref{fig:train_test_loss} for two different initial SGD temperatures; note the higher curvature at lower SGD temperature). Finally, we estimated the curvature of the loss functions along this line by reflecting them about their local minima in the direction from $\theta^\textit{test}_k$ toward $\theta^\textit{tr}_k$, and fitting a quadratic centered at the local minimum (as shown in Figure \ref{fig:train_test_loss}, right pane). (We focus on the line connecting the local minima because all quantities in (\ref{eq:taylor_test_loss}) with $\hat \theta = \theta^\textit{tr}_k$ lie along this line. Letting $\Theta(r)$ denote the line, with $\Theta(0) = \theta^\textit{tr}_k$ and $\Theta(||s_k||) = \theta^\textit{test}_k$, the curvature of the fitted quadratic corresponds to $s_k C_k s_k / \|s_k\|^2$, as shown in (\ref{eq:loss_along_a_line}).) Although we could compute the full Hessian matrix, in these experiments we are mainly interested in the curvature along the line $\Phi(r)$ which can be computed much more cheaply via the 1D projection; further, because of the shapes of the losses we find that the symmetrized quadratic fit better captures the relevant overall curvature of the basin, as discussed in Appendix \ref{append:expt_notes}. The plots in Figures \ref{fig:vgg_shift_curv}, \ref{fig:resnet_nbn_shift_curv} show trends in key quantities (losses and curvature at local minima, shift magnitude, and shift-curvature) as a function of the initial SGD temperature. The results validate (\ref{eq:taylor_test_loss}) (with $\hat \theta = \theta^\textit{tr}_k$) by showing that the test loss at the train minimum is approximately equal to the test loss at the test local minimum $U^\textit{test}_k$ plus the shift-curvature $\half s_k^T C^\textit{test}_k s_k$; empirically $\half s_k^T C^\textit{test}_k s_k$ appears to be more significant than $U^\textit{test}_k$. Further, they show that both the train and test curvatures along the line $\Theta(r)$ decrease with increasing SGD temperature, as does shift-curvature. Interestingly, although our theory makes no predictions about the shift magnitude $\|s_k\|$, in our experiments it increases with temperature for VGG16, but decreases with temperature for Resnet-10 -- but in both cases, the \emph{shift-curvature} still decreases. Also, the test curvature is larger than the train curvature for VGG16 but the opposite is true for Resnet10.

\subsection{A simple two-basin example}
\begin{figure}[t]
\centering
\includegraphics[width=0.47\columnwidth]{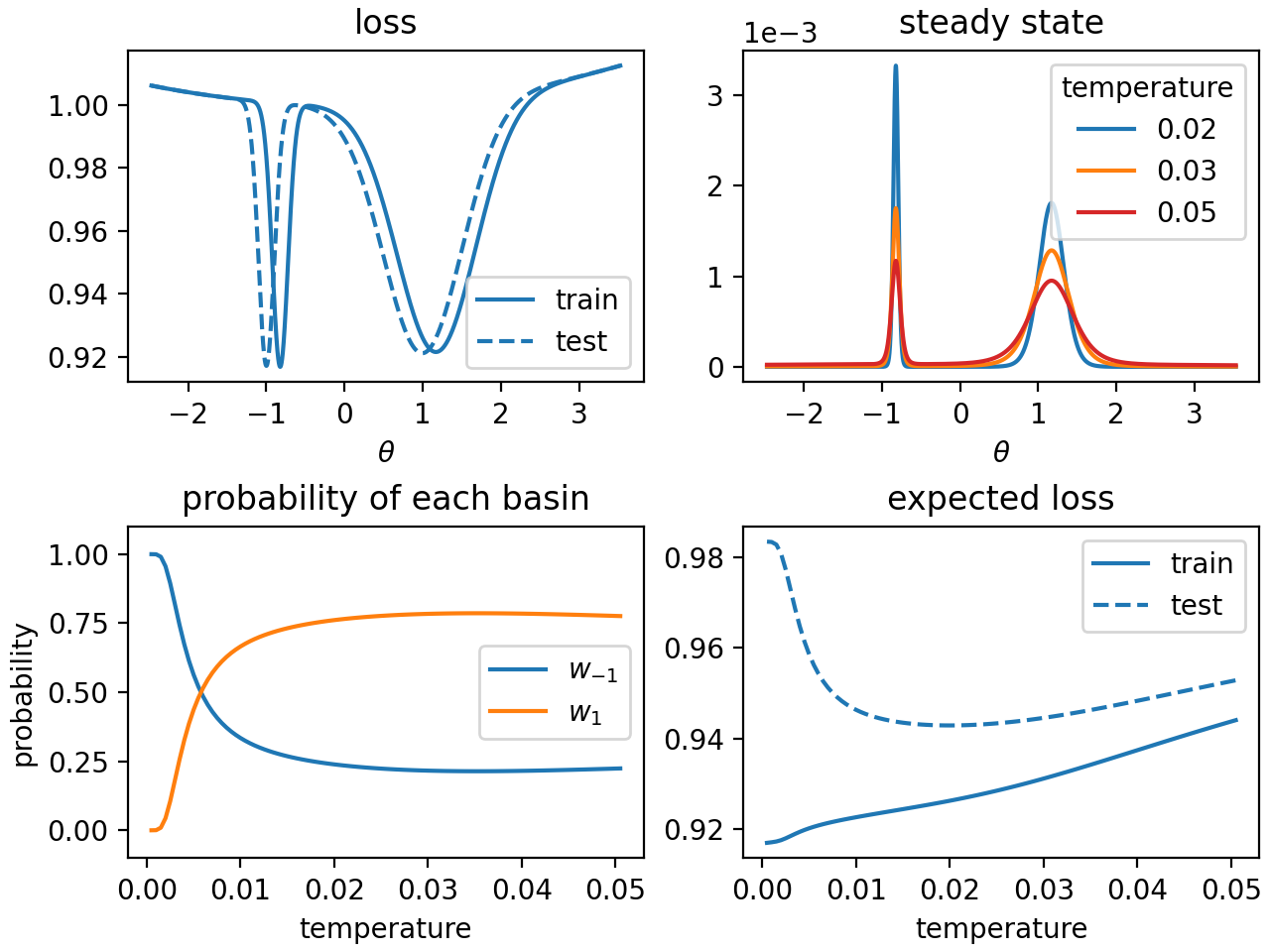}
\quad \quad
\includegraphics[width=0.47\columnwidth]{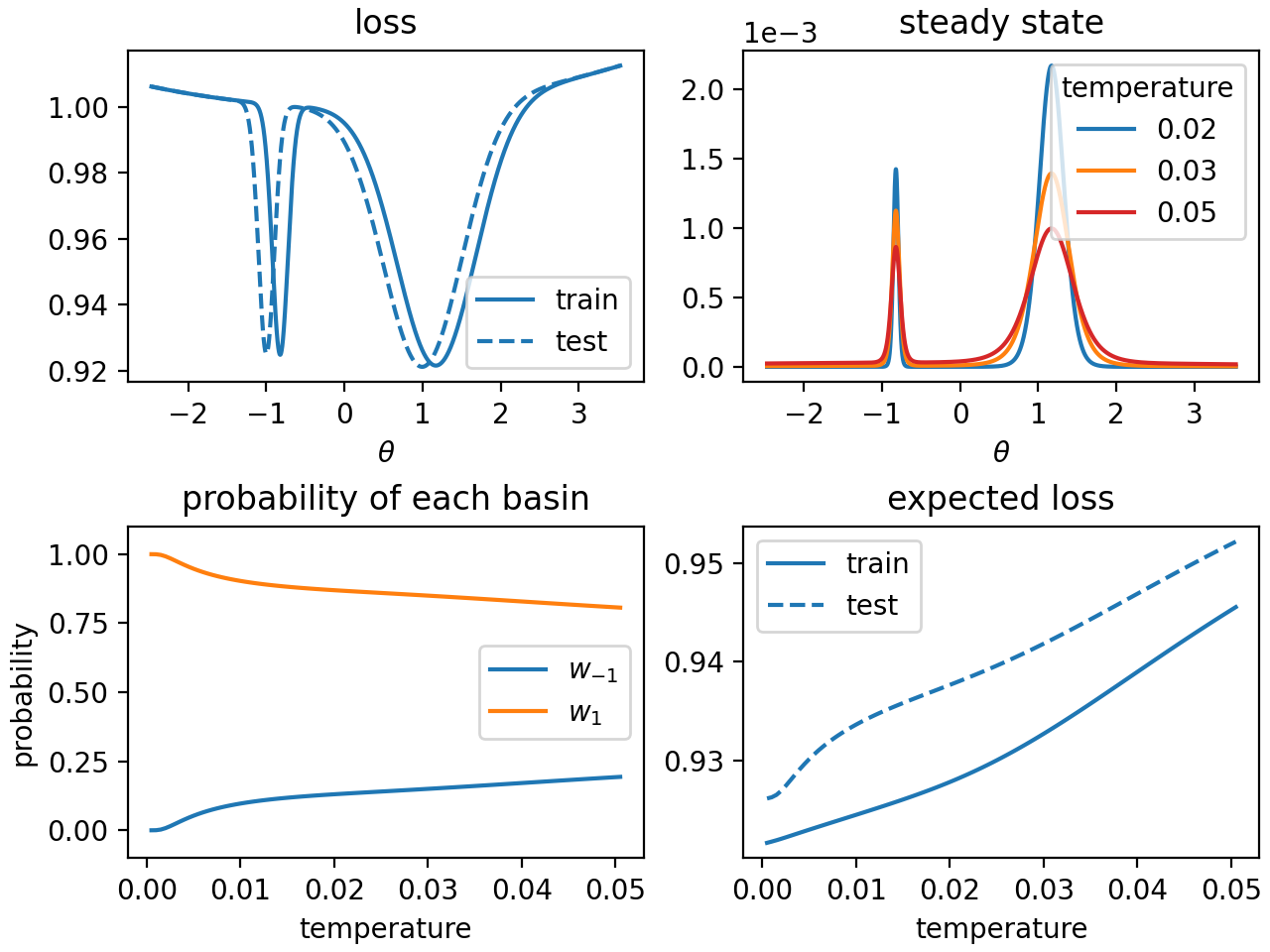}
\caption{
Synthetic two-basin loss with train loss shifted by a constant $s$ relative to test loss, and $D$ constant (so bias is zero). (Left) Minimum at $-1$ is deeper but minimum at $1$ is wider. As temperature increases, the steady-state distribution transfers weight to the wider minimum which has lower shift-curvature, so that test error decreases and training error increases. (Right) Minimum at $1$ is deeper and wider. There is no depth/curvature trade-off to explore, so both training and test error worsen as $T$ increases.
}
\label{fig:2basin_g_A}
\end{figure}
To illustrate the impact of shift and curvature on test loss and the temperature-controlled depth/curvature trade-off, we construct a simplified synthetic optimization problem where $\theta \in \RR,$ and the train and test losses each have two local minima (the data is implicit because we directly define the train and test losses as functions of $\theta$). We construct the train and test losses to be identical except for a constant shift (i.e. $U^\textit{test}(\theta) = U^\textit{tr}(\theta + s)$). We let the noise $D^\textit{tr}$ be a constant as well, so the bias is zero. Thus (\ref{eq:sgdss}) becomes $\rho(\theta) \propto e^{-\frac{2}{T} U^\textit{tr}(\theta)}$, the basin weights become $w_k \propto e^{-\frac{2}{T} U^\textit{tr}_k} (C^\textit{tr}_k)^{-\half}$ and
(\ref{eq:testPerfSGDApprox}) becomes 
\begin{align}
    E_{\theta \sim \rho}[U^{\textit{test}}(\theta)] &\approx \sum_k w_k \big( U^\textit{tr}_k + \half s^T C^\textit{tr}_k s \big),
\end{align}
since $U^\textit{tr}_k = U^{\textit{test}}_k$ and $C^\textit{tr}_k = C^{\textit{test}}_k$. In Figure \ref{fig:2basin_g_A}, the losses each have two local minima with different curvature (hence different shift-curvature, since the shift is constant). On the left, the minimum at $\theta = -1$ (let us assign it $k=0$) is deeper but narrower than the minimum at $\theta = 1$ ($k=1$), so that $U^\textit{tr}_0 < U^\textit{tr}_1$ and $C^\textit{tr}_0 > C^\textit{tr}_1$. At $T=0$, $\rho$ is a Dirac delta at $\argmin (U^\textit{tr})$ and $w_0 = 1$. As $T$ increases, $w_1$ increases, and we get a mixture between the two basins. As $w_1$ increases, the depth term worsens ($w_0 U^\textit{tr}_0 + w_1 U^\textit{tr}_1$ increases), while the shift-curvature terms improves ($\half s^T (w_0 C^\textit{tr}_0 + w_1 C^\textit{tr}_1) s$ decreases). For larger temperatures, the variance and/or bias terms in the loss become large enough to worsen test performance again, so there is a non-zero and finite optimal temperature that minimizes test performance. On the right, we have a similar setup, except that the deeper basin is also wider. In this case, we see no improvement in test loss with increasing temperature, since there is no trade-off to explore between depth and shift-curvature. Appendix \ref{append:more_synth_exp} includes similar one-basin and three-basin cases. In the one-basin case, increasing the temperature can only worsen the expected test loss, since there is no trade-off to explore, and increasing the SGD variance can only hurt both the train and test losses on average. The three basin case is qualitatively similar to the two basin case, but also shows the preference for lower curvature when the two basins are equally deep.

\section{Stochastic Gradient Descent} \label{sec:sgd}

In this section we derive Equations \ref{eq:sgdss}, \ref{eq:weight_rho}, and \ref{eq:testPerfSGDApprox}.
Recall that SGD attempts to minimize the training loss $U^\textit{tr}(\theta)$ through a discrete-time stochastic process with state $\theta_t \in \mathbb{R}^n,$ where $t$ indexes the number of SGD updates:
\begin{align}
 \theta_{t+1} &= \theta_t + \Delta_t, \text{ where }
  \Delta_t = -\frac{\lambda}{B}\sum_{i=1}^B \partial_\theta U(x_i, \theta), \label{eq:SGD} 
\end{align}
where $\lambda$ is the learning rate, $B$ is the minibatch size, and the $x_i$ are i.i.d. samples from the train set that comprise the minibatch. Letting  $T=\lambda/B$, we note that $\Delta_t$ has mean $-\lambda \partial_{\theta} U^\textit{tr}(\theta)$ and covariance $\lambda T D^\textit{tr}(\theta)$ --- we refer to the latter as the SGD noise, and think of the temperature (by analogy with statistical mechanics) as a good summary of noise strength. Typical applications use $\lambda \ll 1$, and $B > 1,$ so that $T \ll 1$, since running SGD with too high a learning rate or temperature results in unstable dynamics (especially early in a run) that often fail to converge, i.e., see \cite{goyal2017accurate}. So we generally assume that $\lambda \ll 1,$ and $T \ll 1$. As shown by \cite{li2017stochastic} and our complementary derivation in Appendix \ref{append:continuous_approx}, under these conditions the probability distribution of parameters $p(\theta, t)$ that SGD produces can be well approximated by the probability distribution $\rho(\theta, t)$ of a continuous-time diffusion process governed by the Fokker-Planck equation
\begin{align}
\partial_t \rho(\theta, t) &= \lambda \partial_\theta \cdot J(\theta, t), \text{ where }  
J(\theta, t) = \partial_{\theta} U^\textit{tr}(\theta) \rho(\theta,t) + \frac{T}{2} \partial_\theta \cdot \big( D^\textit{tr}(\theta) \rho(\theta,t) \big). \label{eq:fp}
\end{align}
$J(\theta, t)$ is called the probability current. Redefining training time through the change of coordinates $t=\lambda t$ has the only effect of removing $\lambda$ from (\ref{eq:fp}), i.e., yielding $\partial_t \rho(\theta, t) = \partial_\theta \cdot J(\theta, t),$ and leaving $T$ as the only equation parameter. (Explicitly, rescaling the temporal axis of SGD runs so that $\lambda_1 n_1 = \lambda_2 n_2$, where $\lambda_1, \lambda_2$ are the learning rates and $n_1, n_2$ are the number of SGD updates for two different runs, should produce similar results as long as the diffusion approximation holds; this can be a useful rule of thumb for adjusting the number of SGD updates as the learning rate is changed.) The fact that $T$ is the only remaining parameter after rescaling time justifies the notion that $T$ is a good summary of SGD noise strength, and means that different SGD runs with the same value of $T$ but different learning rates and minibatch sizes still produce approximately the same steady-state parameter distributions.  The dependence of the SGD distribution on $T$ is in agreement with theory given in \cite{jastrzkebski2017three}, and the empirical results of \cite{goyal2017accurate, he2019control}. We also confirm the $T$ dependence in Figure \ref{fig:vgg_lb_init_hpuh5hk7ja} (Appendix \ref{append:temp_expt}), which shows similar test performance for different minibatch sizes and learning rates as long as $T$ is kept fixed. Finally, \cite{goyal2017accurate} notes that the simple dependence of SGD on $T$ breaks down when the learning rate is sufficiently large even when $T$ is small; a situation where we expect the diffusion approximation to be inaccurate.

With the change of variables $t=\lambda t$ just described, $\partial_t \rho(\theta, t) =  \partial_\theta \cdot J(\theta, t)$ is equivalent to the Langevin (stochastic differential) equation
\begin{align}
    d\theta = - \partial_\theta U^\textit{tr}(\theta)dt + \sqrt{T D^\textit{tr}(\theta)} dW \label{eq:lang}
\end{align} 
that appears in  \cite{li2017stochastic} and \cite{chaudhari2018stochastic} as an approximation of SGD. Fokker-Planck and Langevin equations describe the same underlying continuous-time diffusion process; i.e., see \cite{gardiner2009}.  \cite{li2017stochastic} showed that (\ref{eq:fp}) is an order-1 weak approximation of SGD. In Appendix \ref{append:continuous_approx_general} we offer a complementary informal constructive derivation that shows how to approximate any discrete-time Markov process by a continuous-time diffusion process through the truncation of the infinite Kramers-Moyal expansion, which may be useful for analyzing other algorithms as well. Our approach essentially relies on approximating each SGD step as a sum of i.i.d. infinitesimal increments, and validates a conjecture in \cite{BazantBird2005} that the moments of the continuous-time diffusion approximation are proportional to the cumulants of the discrete-time process.\footnote{Informally, in our derivation we need the probability distribution of $\Delta_t$ to vary slowly with $\theta$, and making $\lambda$ sufficiently small guarantees this, to ensure that infinitesimal increments comprising an SGD step are approximately i.i.d.. We also need $T \ll 1$ to truncate the Kramers-Moyal expansion to second order and obtain (\ref{eq:fp}). The same conditions also appear in \cite{li2017stochastic}: $T \ll 1$ follows from the bound of the accuracy of the order-1 weak approximation, which is a constant times the learning rate with batch size assumed constant. That the probability of the updates varies slowly with $\theta$ appears formally there as a Lipschitz and growth condition on the loss.} For the rest of this paper, we assume that (\ref{eq:fp}) is a good approximation to SGD, and study the resulting stationary distribution. 
Next, we assume that the process described by (\ref{eq:fp}) is ergodic, and seek its unique steady-state solution, i.e., the distribution $\rho(\theta)$ such that $\partial_t \rho(\theta, t) = 0$. We can follow \cite{gardiner2009}
and seek a solution where $J(\theta, t)=0.$ Appendix \ref{append:steady-state} shows that some algebra starting from $J(\theta, t)=0$, assuming that the integrand of the line integral in (\ref{eq:sgdss}) is a gradient, yields the steady-state distribution in (\ref{eq:sgdss}). This approach needs an invertible $D^\textit{tr}(\theta),$ which is only possible in the underparametrized case. Section \ref{sec:overparam} describes what happens in the overparametrized case.

To approximate the expected SGD test loss (\ref{eq:testPerfSGDApprox}), we approximate the SGD steady-state distribution as a mixture of Gaussians (similar to \cite{jastrzkebski2017three}). Since (\ref{eq:sgdss}) has the form $\rho(\theta) \propto e^{-\frac{2}{T} v(\theta)}$, it is reasonable to expect that that when $T\ll1,$ $\rho$ is multimodal with peaks at the local minima of the effective potential $v(\theta)$. So we make a Laplace approximation, i.e., we approximate $v(\theta)$ by a quadratic function in a neighborhood of any local minimum to obtain a local Gaussian approximation. Combining the local approximation at every local minimum yields a weighted Gaussian mixture approximation of $\rho(\theta)$, with weights given by (\ref{eq:weight_rho}), and bias and variance both $O(T)$. Details are in Appendix \ref{append:sgd_ss_gmm}. (Note that the bias is linear in $T$ but nonzero, confirming a result in \cite{chaudhari2018stochastic} that the critical points of $v$ differ linearly-in-$T$ from those of $U^\textit{tr}$.)
Substituting these results into the approximate test loss (\ref{eq:approxAvgTestLoss}) and keeping only leading terms in $T$ results in the approximate SGD test performance in (\ref{eq:testPerfSGDApprox}).  The validity of the Gaussian mixture approximation of (\ref{eq:sgdss}) depends on the local quadratic approximation of $v$, which we explore in our experiments with quadratic fits to the train loss, to which $v$ is related. 

\section{More Parameters Than Datapoints} \label{sec:overparam}

We now consider the overparametrized situation ($p \geq N_\textit{tr}$ and/or $ p \geq N_\textit{test}$, recalling that $p$ denotes the number of model parameters and $N_\textit{tr/test}$ the number train/test model parameters). As discussed in Appendix \ref{append:overparametrization}, the overparametrized case is more complex because the per-sample gradients of the train set now do not span $\RR^p,$ but rather a subspace of dimension no larger than $N_\textit{tr}$, denoted $\mathcal{G}_\textit{tr}$, which in general depends on $\theta$ (similarly, there is a subspace $\mathcal{G}_\textit{test}$ of dimension $N_\textit{test}$ for test). This has many implications. One is that the losses no longer necessarily have strict local minima, which means we need to generalize the definition of the shift and the Taylor approximation (\ref{eq:taylor_test_loss}). Another is that the average training gradient $\partial_\theta U^\textit{tr}(\theta)$ and the range of $D^\textit{tr}(\theta)$ both lie in $\mathcal{G}_\textit{tr}$, which implies both that that the SGD updates $\Delta_t$ in (\ref{eq:SGD}) lie in $\mathcal{G}_\textit{tr}$, and that $D^\textit{tr}(\theta)$ is no longer invertible, invalidating (\ref{eq:sgdss}). To help resolve these issues, we will need to add $\ell_2$-regularization to the train loss (as is common in practice), and also to modify SGD to include isotropic noise. With these changes, we can obtain analogs of our main results.

In Appendix \ref{append:taylor_overparam}, we generalize (\ref{eq:taylor_test_loss}) to the overparametrized case. We first add $\ell_2$-regularization to the train loss:
$U^{\textit{tr}}(\theta) = E_{x \sim \textit{tr}}[U(x, \theta)] + \half \alpha \theta^T \theta,$
which makes the train local minima strict (so that the bias $\hat b$ in (\ref{eq:taylor_test_loss}) makes sense); however, the test loss is unregularized so its local minima are non-strict. We find that (\ref{eq:taylor_test_loss}) still holds if we generalize the definition of the shift to
\begin{align} 
    s_k &= \text{Proj}_{\mathcal{G}_\textit{test}} \big( \theta^{\textit{test}}_k - \theta^\textit{tr}_k \big),
\end{align}
where $\text{Proj}_{\mathcal{G}_\textit{test}}$ is the projector onto the subspace spanned by the per-test-sample gradients.

In order to generalize Equation \ref{eq:sgdss}, we modify SGD to include isotropic noise in addition to the $\ell_2$-regularization of the train loss, as follows:
\begin{align}
 \theta_{t+1} &= \theta_t + \Delta_t, \text{ where }
  \Delta_t = -\frac{\lambda}{B}\sum_{i=1}^B \partial_\theta U(x_i, \theta) - \alpha \theta + \sqrt{\lambda T} \beta w_t, \label{eq:SGDMod} 
\end{align}
where $\alpha, \beta$ are scalars, and where $w_t$ is zero-mean Gaussian noise in $\RR^p$ with covariance equal to the identity, that is independent of everything else. The isotropic Gaussian noise ensures that $D^\textit{tr}(\theta)$ is invertible, while $\ell_2$-regularization controls the part of the drift in $\RR^p$ but not in $\mathcal{G}_\textit{tr}$, as discussed in Appendix \ref{append:ss_overparam}. The diffusion process approximation of the modified SGD is then exactly like the one for the underparametrized case after making the substitutions $D^\textit{tr}(\theta) + \beta^2 I$ in place of $D^\textit{tr}(\theta)$, and $\partial_\theta U^\textit{tr}( \theta) + \alpha \theta$ in place of  $\partial_\theta U^\textit{tr}( \theta).$ With these substitutions, the steady-state distribution still has the form in Equation \ref{eq:sgdss}, and its Gaussian mixture approximation still has the form in Equation \ref{eq:testPerfSGDApprox}.

\section{Other Related Work}
There are many works aimed at understanding generalization in the context of SGD, and the interplay between SGD noise, curvature, and test performance. In addition to the most relevant works already discussed: \cite{smith2017bayesian, he2019control, mandt2017stochastic, ahn2012bayesian} take a Bayesian perspective, \cite{he2019asymmetric} study asymmetrical valleys, \cite{belkin2019reconciling} focus on model capacity, \cite{martin2018implicit} apply random matrix theory, \cite{corneanu2020computing} propose persistent topology measures, \cite{russo2016controlling, xu2017information} provide information-theoretic bounds, \cite{smith2021origin} perform error analysis relative to gradient flow, \cite{lee2017gp, khan2019approximate} connect to Gaussian processes, \cite{wu2019multiplicative} analyze multiplicative noise, and \cite{zhu2018anisotropic} study `escaping efficiency'. \cite{he2019asymmetric} show that SGD is biased toward the flatter side of asymmetrical loss valleys, and such a bias can improve generalization (offering additional insight into the bias term in (\ref{eq:taylor_test_loss}) which we identify but do not focus on).
Although \cite{smith2017bayesian, he2019control, mandt2017stochastic} make important connections between generalization and curvature, none provide an explicit expression for test loss in terms of curvature. Also, their analyses are focused around a single local minimum, rather than comparing different local minima as we do in this paper. \cite{mandt2017stochastic} show that SGD can perform approximate Bayesian inference by studying a diffusion approximation of SGD and its steady-state, but assuming a single, quadratic minimum, and constant isotropic noise; assumptions we relax in this paper.
\cite{he2019control} derive a PAC-Bayes bound on the generalization gap at a given local minimum, which shows that wider minima have a lower upper-bound on risk; in contrast, we make a direct approximation showing that wider minima have lower test loss. They also show that the bound on the risk can increase as log(1/T) for a particular local minimum, provided the model is large relative to the $T$ and curvature; this is quite different from our analysis, which involves analyzing the role of $T$ in modulating the SGD steady-state distribution's preference for different local minima (and with no assumptions on model size). 
\cite{smith2017bayesian} connect Bayesian evidence to curvature at a single local minimum, and empirically suggest evidence as proxy for test loss (by contrast we connect test loss directly to curvature). They also show that the steady state of a Langevin equation with constant isotropic noise is proportional to evidence, and by analogy surmise that SGD noise drives the steady-state toward wider minima; we agree with their intuition, but we allow for non-constant non-isotropic noise, and use a Gaussian mixture approximation to directly show the preference for wider minima. Finally, there is a long history of using statistical mechanics techniques to study machine learning, as we do in this work. For example, \cite{seung1992statistical} prove among many other results that the generalization gap is positive; however, like many works both before and after, they assume constant and isotropic noise. Our work motivates updating these earlier studies to the algorithms currently used in practice, e.g., by using the parameter distribution in (\ref{eq:sgdss}).

\section{Conclusion and Discussion}

This paper contributes to longstanding debates about whether curvature can predict generalization, and how and whether SGD minimizes curvature. First, we show that curvature harms test performance through three mechanisms: the shift-curvature, bias-curvature, and the covariance-curvature; we believe the first two to be new. We address the concern of \cite{dinh2017sharp} by showing that all three are reparametrization-invariant although curvature is not. Our main focus is the shift-curvature, or curvature along the line connecting train and test local minima; although the shift is unknown at training time, the shift-curvature still shows that \emph{any} directions with high curvature can potentially harm test performance if they happen to align with the shift. Second, we derive a new and explicit SGD steady-state distribution showing that SGD optimizes an effective potential related to but different from train loss, and show that SGD noise mediates a trade-off between the depth and curvature of this effective potential. Our steady-state solution removes assumptions in earlier works of constant, isotropic gradient covariance, and treats the overparametrized case with care. Third, we combine our test performance analysis with the SGD steady-state to show that for small SGD noise, shift-curvature may be the dominant mechanism. Our experiments validate our approximations of test loss, show that that shift-curvature is indeed a major contributor to test performance, and show that SGD with higher noise chooses local minima with lower curvature and lower shift-curvature.

However, many avenues for future work still remain.
First of all, we have made substantial, but not complete, progress in the overparametrized case. One of the main challenges is directions of near-zero curvature (a concern raised by \cite{draxler2018essentially, sagun2018empirical} and others). On the positive side, overparametrization poses no problem for our test performance approximation (\ref{eq:taylor_test_loss}) or shift-curvature concept (since zero-curvature directions simply have no affect on either local test performance or shift-curvature). Furthermore, our SGD results eliminate constant, isotropic assumptions that appear in earlier works, clarify the challenges in the overparametrized case (such as the rank-deficient gradient covariance with varying nullspace), and still obtain the steady-state solution (\ref{eq:sgdss}) of a slightly modified problem. However, the Gaussian mixture approximation (\ref{eq:weight_rho}), which highlights SGD's low-curvature preference, is less clear in the presence of near-zero-curvature directions. Therefore, further work is needed to clarify how near-zero-curvature directions may affect the unmodified SGD steady-state as well as its preference for low-curvature regions. (A possibly-useful intuition is that, locally, SGD should force all near-zero-curvature directions to zero, making them essentially irrelevant; however, there are difficulties in making this precise, as discussed in Appendix \ref{append:overparametrization}.)
Another different limitation of our work is the assumption that SGD reaches steady state (which enables us to obtain and study en explicit parameter distribution), which may be too slow to accomplish and difficult to verify in practice. We also assume a constant learning rate, although practitioners often use learning-rate schedules. To better understand these situations, further work is needed to study the dynamics of SGD and its impact on generalization, possibly using the diffusion approximation (\ref{eq:fp}). Other popular learning algorithms, like ADAM and SGD with momentum, could be studied using the techniques presented here. The diffusion approximation in (\ref{eq:fp}) might yield additional insights into SGD. 
For example, the so-called fluctuation-dissipation relations derived in \cite{yaida2018fluctuation} from the exact SGD process in (\ref{eq:SGD}) follow more directly from (\ref{eq:fp}) by considering the time derivative of the average of any function of the parameters, and applying this to the parameters themselves, the parameter fluctuations, and the train loss (although the relation for train loss is quadratic in $T$, and deriving it via (\ref{eq:fp}) only recovers the constant and linear in $T$ terms). Studying the time derivative of other functions of the parameters may yield additional relations between key quantities. Third, our setup could be a helpful starting point for investigating model-wise double descent (\cite{belkin2019reconciling}), by studying how the stationary distribution (\ref{eq:sgdss}) and test performance change as model complexity increases (perhaps asymptotically via random matrix theory). Finally, our results suggest that further work on models and algorithms that promote low-curvature solutions, such as the algorithm recently proposed in \cite{orvieto2022}, may have great practical value.

\bibliography{sgdbib}

\begin{thebibliography}{48}
\providecommand{\natexlab}[1]{#1}
\providecommand{\url}[1]{\texttt{#1}}
\expandafter\ifx\csname urlstyle\endcsname\relax
  \providecommand{\doi}[1]{doi: #1}\else
  \providecommand{\doi}{doi: \begingroup \urlstyle{rm}\Url}\fi

\bibitem[Ahn et~al.(2012)Ahn, Korattikara, and Welling]{ahn2012bayesian}
Sungjin Ahn, Anoop Korattikara, and Max Welling.
\newblock Bayesian posterior sampling via stochastic gradient fisher scoring.
\newblock \emph{arXiv preprint arXiv:1206.6380}, 2012.

\bibitem[Bazant(2005)]{BazantBird2005}
Martin Bazant.
\newblock Lecture 9: Kramers-moyall cumulant expansion, scribed by jacy bird.
\newblock \url{https://math.mit.edu/classes/18.366/lec05/lec09.pdf}, March
  2005.

\bibitem[Belkin(2021)]{belkin2021fit}
Mikhail Belkin.
\newblock Fit without fear: remarkable mathematical phenomena of deep learning
  through the prism of interpolation.
\newblock \emph{arXiv preprint arXiv:2105.14368}, 2021.

\bibitem[Belkin et~al.(2019)Belkin, Hsu, Ma, and Mandal]{belkin2019reconciling}
Mikhail Belkin, Daniel Hsu, Siyuan Ma, and Soumik Mandal.
\newblock Reconciling modern machine-learning practice and the classical
  bias--variance trade-off.
\newblock \emph{Proceedings of the National Academy of Sciences}, 116\penalty0
  (32):\penalty0 15849--15854, 2019.

\bibitem[Buntine(1991)]{buntine1991bayesian}
Wray~L Buntine.
\newblock Bayesian backpropagation.
\newblock \emph{Complex systems}, 5:\penalty0 603--643, 1991.

\bibitem[Chaudhari and Soatto(2018)]{chaudhari2018stochastic}
Pratik Chaudhari and Stefano Soatto.
\newblock Stochastic gradient descent performs variational inference, converges
  to limit cycles for deep networks.
\newblock In \emph{2018 Information Theory and Applications Workshop (ITA)},
  pages 1--10. IEEE, 2018.

\bibitem[Comon(1994)]{comon1994tensor}
Pierre Comon.
\newblock Tensor diagonalization, a useful tool in signal processing.
\newblock \emph{IFAC Proceedings Volumes}, 27\penalty0 (8):\penalty0 77--82,
  1994.

\bibitem[Corneanu et~al.(2020)Corneanu, Escalera, and
  Martinez]{corneanu2020computing}
Ciprian~A Corneanu, Sergio Escalera, and Aleix~M Martinez.
\newblock Computing the testing error without a testing set.
\newblock In \emph{Proceedings of the IEEE/CVF Conference on Computer Vision
  and Pattern Recognition}, pages 2677--2685, 2020.

\bibitem[Dinh et~al.(2017)Dinh, Pascanu, Bengio, and Bengio]{dinh2017sharp}
Laurent Dinh, Razvan Pascanu, Samy Bengio, and Yoshua Bengio.
\newblock Sharp minima can generalize for deep nets.
\newblock In \emph{Proceedings of the 34th International Conference on Machine
  Learning-Volume 70}, pages 1019--1028. JMLR. org, 2017.

\bibitem[Draxler et~al.(2018)Draxler, Veschgini, Salmhofer, and
  Hamprecht]{draxler2018essentially}
Felix Draxler, Kambis Veschgini, Manfred Salmhofer, and Fred Hamprecht.
\newblock Essentially no barriers in neural network energy landscape.
\newblock In \emph{International conference on machine learning}, pages
  1309--1318. PMLR, 2018.

\bibitem[Gardiner(2009)]{gardiner2009}
Crispin Gardiner.
\newblock \emph{Elements of Stochastic Methods}, volume~4.
\newblock Springer Berlin, 2009.

\bibitem[Golmant et~al.(2018)Golmant, Vemuri, Yao, Feinberg, Gholami, Rothauge,
  Mahoney, and Gonzalez]{golmant2018computational}
Noah Golmant, Nikita Vemuri, Zhewei Yao, Vladimir Feinberg, Amir Gholami, Kai
  Rothauge, Michael~W Mahoney, and Joseph Gonzalez.
\newblock On the computational inefficiency of large batch sizes for stochastic
  gradient descent.
\newblock \emph{arXiv preprint arXiv:1811.12941}, 2018.

\bibitem[Goyal et~al.(2017)Goyal, Doll{\'a}r, Girshick, Noordhuis, Wesolowski,
  Kyrola, Tulloch, Jia, and He]{goyal2017accurate}
Priya Goyal, Piotr Doll{\'a}r, Ross Girshick, Pieter Noordhuis, Lukasz
  Wesolowski, Aapo Kyrola, Andrew Tulloch, Yangqing Jia, and Kaiming He.
\newblock Accurate, large minibatch sgd: Training imagenet in 1 hour.
\newblock \emph{arXiv preprint arXiv:1706.02677}, 2017.

\bibitem[He et~al.(2019{\natexlab{a}})He, Liu, and Tao]{he2019control}
Fengxiang He, Tongliang Liu, and Dacheng Tao.
\newblock Control batch size and learning rate to generalize well: Theoretical
  and empirical evidence.
\newblock \emph{Advances in Neural Information Processing Systems},
  32:\penalty0 1143--1152, 2019{\natexlab{a}}.

\bibitem[He et~al.(2019{\natexlab{b}})He, Huang, and Yuan]{he2019asymmetric}
Haowei He, Gao Huang, and Yang Yuan.
\newblock Asymmetric valleys: Beyond sharp and flat local minima.
\newblock \emph{Advances in neural information processing systems}, 32,
  2019{\natexlab{b}}.

\bibitem[He et~al.(2016)He, Zhang, Ren, and Sun]{he2016deep}
Kaiming He, Xiangyu Zhang, Shaoqing Ren, and Jian Sun.
\newblock Deep residual learning for image recognition.
\newblock In \emph{Proceedings of the IEEE conference on computer vision and
  pattern recognition}, pages 770--778, 2016.

\bibitem[Hochreiter and Schmidhuber(1997)]{hochreiter1997flat}
Sepp Hochreiter and J{\"u}rgen Schmidhuber.
\newblock Flat minima.
\newblock \emph{Neural computation}, 9\penalty0 (1):\penalty0 1--42, 1997.

\bibitem[Hoffer et~al.(2017)Hoffer, Hubara, and Soudry]{hoffer2017trainlonger}
Elad Hoffer, Itay Hubara, and Daniel Soudry.
\newblock Train longer, generalize better: closing the generalization gap in
  large batch training of neural networks.
\newblock In \emph{Advances in Neural Information Processing Systems}, pages
  1731--1741, 2017.

\bibitem[Jastrzkebski et~al.(2017)Jastrzkebski, Kenton, Arpit, Ballas, Fischer,
  Bengio, and Storkey]{jastrzkebski2017three}
Stanislaw Jastrzkebski, Zachary Kenton, Devansh Arpit, Nicolas Ballas, Asja
  Fischer, Yoshua Bengio, and Amos Storkey.
\newblock Three factors influencing minima in sgd.
\newblock \emph{arXiv preprint arXiv:1711.04623}, 2017.

\bibitem[Jiang et~al.(2019)Jiang, Neyshabur, Mobahi, Krishnan, and
  Bengio]{jiang2019fantastic}
Yiding Jiang, Behnam Neyshabur, Hossein Mobahi, Dilip Krishnan, and Samy
  Bengio.
\newblock Fantastic generalization measures and where to find them.
\newblock \emph{arXiv preprint arXiv:1912.02178}, 2019.

\bibitem[Keskar et~al.(2016)Keskar, Mudigere, Nocedal, Smelyanskiy, and
  Tang]{keskar2016large}
Nitish~Shirish Keskar, Dheevatsa Mudigere, Jorge Nocedal, Mikhail Smelyanskiy,
  and Ping Tak~Peter Tang.
\newblock On large-batch training for deep learning: Generalization gap and
  sharp minima.
\newblock \emph{arXiv preprint arXiv:1609.04836}, 2016.

\bibitem[Khan et~al.(2019)Khan, Immer, Abedi, and Korzepa]{khan2019approximate}
Mohammad~Emtiyaz Khan, Alexander Immer, Ehsan Abedi, and Maciej Korzepa.
\newblock Approximate inference turns deep networks into gaussian processes.
\newblock \emph{arXiv preprint arXiv:1906.01930}, 2019.

\bibitem[Krizhevsky et~al.(2009)Krizhevsky, Hinton,
  et~al.]{krizhevsky2009learning}
Alex Krizhevsky, Geoffrey Hinton, et~al.
\newblock Learning multiple layers of features from tiny images, 2009.

\bibitem[Lee et~al.(2017)Lee, Bahri, Novak, Schoenholz, Pennington, and
  Sohl-Dickstein]{lee2017gp}
Jaehoon Lee, Yasaman Bahri, Roman Novak, Samuel~S Schoenholz, Jeffrey
  Pennington, and Jascha Sohl-Dickstein.
\newblock Deep neural networks as gaussian processes.
\newblock \emph{arXiv preprint arXiv:1711.00165}, 2017.

\bibitem[Li et~al.(2017{\natexlab{a}})Li, Xu, Taylor, Studer, and
  Goldstein]{li2017visualizing}
Hao Li, Zheng Xu, Gavin Taylor, Christoph Studer, and Tom Goldstein.
\newblock Visualizing the loss landscape of neural nets.
\newblock \emph{arXiv preprint arXiv:1712.09913}, 2017{\natexlab{a}}.

\bibitem[Li et~al.(2017{\natexlab{b}})Li, Tai, et~al.]{li2017stochastic}
Qianxiao Li, Cheng Tai, et~al.
\newblock Stochastic modified equations and adaptive stochastic gradient
  algorithms.
\newblock In \emph{Proceedings of the 34th International Conference on Machine
  Learning-Volume 70}, pages 2101--2110. JMLR. org, 2017{\natexlab{b}}.

\bibitem[Li et~al.(2020)Li, Gu, Zhou, Chen, and Banerjee]{li2020hessian}
Xinyan Li, Qilong Gu, Yingxue Zhou, Tiancong Chen, and Arindam Banerjee.
\newblock Hessian based analysis of sgd for deep nets: Dynamics and
  generalization.
\newblock In \emph{Proceedings of the 2020 SIAM International Conference on
  Data Mining}, pages 190--198. SIAM, 2020.

\bibitem[Liu and Deng(2015)]{liu2015very}
Shuying Liu and Weihong Deng.
\newblock Very deep convolutional neural network based image classification
  using small training sample size.
\newblock In \emph{2015 3rd IAPR Asian conference on pattern recognition
  (ACPR)}, pages 730--734. IEEE, 2015.

\bibitem[Mandt et~al.(2017)Mandt, Hoffman, and Blei]{mandt2017stochastic}
Stephan Mandt, Matthew~D Hoffman, and David~M Blei.
\newblock Stochastic gradient descent as approximate bayesian inference.
\newblock \emph{The Journal of Machine Learning Research}, 18\penalty0
  (1):\penalty0 4873--4907, 2017.

\bibitem[Martin and Mahoney(2018)]{martin2018implicit}
Charles~H Martin and Michael~W Mahoney.
\newblock Implicit self-regularization in deep neural networks: Evidence from
  random matrix theory and implications for learning.
\newblock \emph{arXiv preprint arXiv:1810.01075}, 2018.

\bibitem[McCandlish et~al.(2018)McCandlish, Kaplan, Amodei, and
  Team]{mccandlish2018empirical}
Sam McCandlish, Jared Kaplan, Dario Amodei, and OpenAI~Dota Team.
\newblock An empirical model of large-batch training.
\newblock \emph{arXiv preprint arXiv:1812.06162}, 2018.

\bibitem[Orvieto et~al.(2022)Orvieto, Kersting, Proske, Bach, and
  Lucchi]{orvieto2022}
Antonio Orvieto, Hans Kersting, Frank Proske, Francis Bach, and Aurelien
  Lucchi.
\newblock Anticorrelated noise injection for improved generalization.
\newblock \emph{arXiv preprint arXiv:2202.02831}, 2022.

\bibitem[Qi(2005)]{qi2005eigenvalues}
Liqun Qi.
\newblock Eigenvalues of a real supersymmetric tensor.
\newblock \emph{Journal of Symbolic Computation}, 40\penalty0 (6):\penalty0
  1302--1324, 2005.

\bibitem[Russo and Zou(2016)]{russo2016controlling}
Daniel Russo and James Zou.
\newblock Controlling bias in adaptive data analysis using information theory.
\newblock In \emph{Artificial Intelligence and Statistics}, pages 1232--1240.
  PMLR, 2016.

\bibitem[Sagun et~al.(2018)Sagun, Evci, G{\"u}ney, Dauphin, and
  Bottou]{sagun2018empirical}
L~Sagun, U~Evci, U~G{\"u}ney, Y~Dauphin, and L~Bottou.
\newblock Empirical analysis of the hessian of over-parametrized neural
  networks, iclr.
\newblock \emph{arXiv preprint arXiv:1706.04454}, 2018.

\bibitem[Seung et~al.(1992)Seung, Sompolinsky, and
  Tishby]{seung1992statistical}
Hyunjune~Sebastian Seung, Haim Sompolinsky, and Naftali Tishby.
\newblock Statistical mechanics of learning from examples.
\newblock \emph{Physical review A}, 45\penalty0 (8):\penalty0 6056, 1992.

\bibitem[Shallue et~al.(2018)Shallue, Lee, Antognini, Sohl-Dickstein, Frostig,
  and Dahl]{shallue2018measuring}
Christopher~J Shallue, Jaehoon Lee, Joe Antognini, Jascha Sohl-Dickstein, Roy
  Frostig, and George~E Dahl.
\newblock Measuring the effects of data parallelism on neural network training.
\newblock \emph{arXiv preprint arXiv:1811.03600}, 2018.

\bibitem[Shon(2021)]{shon2021resnet}
Hyounguk Shon.
\newblock Simple cifar10 resnet in jax.
\newblock \url{https://github.com/hushon/JAX-ResNet-CIFAR10}, 2021.

\bibitem[Simonyan and Zisserman(2014)]{simonyan2014very}
Karen Simonyan and Andrew Zisserman.
\newblock Very deep convolutional networks for large-scale image recognition.
\newblock \emph{arXiv preprint arXiv:1409.1556}, 2014.

\bibitem[Smith et~al.(2020)Smith, Elsen, and De]{smith2020generalization}
Samuel Smith, Erich Elsen, and Soham De.
\newblock On the generalization benefit of noise in stochastic gradient
  descent.
\newblock In \emph{International Conference on Machine Learning}, pages
  9058--9067. PMLR, 2020.

\bibitem[Smith and Le(2017)]{smith2017bayesian}
Samuel~L Smith and Quoc~V Le.
\newblock A bayesian perspective on generalization and stochastic gradient
  descent.
\newblock \emph{arXiv preprint arXiv:1710.06451}, 2017.

\bibitem[Smith et~al.(2021)Smith, Dherin, Barrett, and De]{smith2021origin}
Samuel~L Smith, Benoit Dherin, David~GT Barrett, and Soham De.
\newblock On the origin of implicit regularization in stochastic gradient
  descent.
\newblock \emph{arXiv preprint arXiv:2101.12176}, 2021.

\bibitem[Wu et~al.(2019)Wu, Hu, Xiong, Huan, and Zhu]{wu2019multiplicative}
Jingfeng Wu, Wenqing Hu, Haoyi Xiong, Jun Huan, and Zhanxing Zhu.
\newblock The multiplicative noise in stochastic gradient descent:
  Data-dependent regularization, continuous and discrete approximation.
\newblock \emph{arXiv preprint arXiv:1906.07405}, 2019.

\bibitem[Xu and Raginsky(2017)]{xu2017information}
Aolin Xu and Maxim Raginsky.
\newblock Information-theoretic analysis of generalization capability of
  learning algorithms.
\newblock \emph{arXiv preprint arXiv:1705.07809}, 2017.

\bibitem[Yaida(2018)]{yaida2018fluctuation}
Sho Yaida.
\newblock Fluctuation-dissipation relations for stochastic gradient descent.
\newblock \emph{arXiv preprint arXiv:1810.00004}, 2018.

\bibitem[Yoshida(2021)]{yoshida2021vgg}
Davis Yoshida.
\newblock Vgg16 in haiku.
\newblock \url{https://github.com/davisyoshida/vgg16-haiku}, 2021.

\bibitem[You et~al.(2017)You, Gitman, and Ginsburg]{you2017scaling}
Yang You, Igor Gitman, and Boris Ginsburg.
\newblock Scaling sgd batch size to 32k for imagenet training.
\newblock \emph{arXiv preprint arXiv:1708.03888}, 6\penalty0 (12):\penalty0 6,
  2017.

\bibitem[Zhu et~al.(2018)Zhu, Wu, Yu, Wu, and Ma]{zhu2018anisotropic}
Zhanxing Zhu, Jingfeng Wu, Bing Yu, Lei Wu, and Jinwen Ma.
\newblock The anisotropic noise in stochastic gradient descent: Its behavior of
  escaping from minima and regularization effects.
\newblock \emph{arXiv preprint arXiv:1803.00195}, 2018.

\end{thebibliography}


\acks{We would like to thank Luca Zappella and Santiago Akle for inspiring us to study this problem; Vimal Thilak for assistance with compute infrastructure, Josh Susskind, Moises Goldszmidt, and Omid Saremi for helpful discussion and suggestions on the draft; Rudolph van der Merwe and John Giannandrea for their support; and our Ph.D. advisors Wing H. Wong and George C. Verghese for first introducing us to the techniques behind this work.}


\newpage

\appendix

\section{Continuous-Time Approximation of SGD}
\label{append:continuous_approx}

\subsection{Continuous-Time Approximation of A Discrete-Time Continuous-Space Markov Process} \label{append:continuous_approx_general}

Our goal in this section is to construct a continuous-time Markov process that approximates a discrete-time Markov process. Later we will specialize to SGD, but for now we consider any discrete-time stochastic process with state $\theta_j \in \RR^p$ (where $j$ indexes time), that evolves according to 
$$ \theta_{j+1} = \theta_j + \Delta_j, $$
where $\Delta_j \in \RR^p$ is an arbitrary function of $\theta_j$. The SGD update is a special case when $\Delta$ is given by (\ref{eq:SGD}). (Note the different notation from the main text: $j$ rather than $t$ is the discrete time index here, so that we can use $t$ to represent continuous time.) We wish to construct a continuous-time Markov process $\tilde{\theta}(t) \in \RR^p$, with probability distribution $\rho(\theta, t) = P(\tilde{\theta}(t) = \theta)$, that approximates $\theta_j$ in the following sense. We assume the discrete-time updates of $\theta_j$ occur every $\tau > 0$ continuous-time units of $\tilde{\theta}$, where $\tau$ is arbitrary. We want $\tilde{\theta}(t)$ such that if $\rho(\theta, (j-1) \tau) = P(\theta_{j-1}=\theta)$ then
\begin{gather}
\rho(\theta, j\tau) \equiv P(\tilde{\theta}(j\tau) = \theta) \approx P(\theta_{j}=\theta). \label{eq:approxP}
\end{gather}

We will construct $\tilde{\theta}$ via its Kramers-Moyal expansion, which is a differential equation that describes the evolution of the probability distribution of $\tilde{\theta}$. Truncating the Kramers-Moyal expansion to second order yields a Fokker-Planck approximation of the discrete-time stochastic process (which we will later specialize to the case of SGD). The Kramers-Moyal equation depends on the moments of the infinitesimal increments of the continuous process (i.e. $\tilde{\theta}(t') - \tilde{\theta}(t)$, for $t' - t$ infinitesimally small), so our basic strategy is to construct these moments in such a way that the continuous-time update over time interval $\tau$ matches the discrete-time update (we will find that defining the continuous moments proportional to the discrete cumulants achieves this). 

We state the final results here, and justify them in the next sections. First we need to introduce some notation. We adopt multi-index notation to streamline the exposition, so for example if $x \in \RR^p$ then $x^\gamma \equiv x_0^{\gamma_0} \ldots x_p^{\gamma_p}$, where $\gamma \in \NN^p$. Dropping the time index $j$ from the discrete-time update $\Delta$ for now, we denote the moments and cumulants of $\Delta$ (which play an important role) by $m_\gamma(\theta) =  E[\Delta^{\gamma}]$ and $\kappa_\gamma(\theta) = \log m_\gamma(\theta)$, respectively, with $\gamma \in \NN^p$ (using multi-index notation for $\Delta^{\gamma}$). Next, we define the increment of the continuous-time process $\tilde{\theta}$ for any two times $t' \geq t$ as the random variable $\tilde{\Delta}(dt) \equiv \tilde{\Delta}(t'-t) = \tilde{\theta}(t') - \tilde{\theta}(t).$ (Note that the Markov property implies that the increment only depends on the time difference $dt = t'-t$ when the value of $\tilde{\theta}(t)$ is known.) 

With this notation, we can present our main results. In order to specify the process $\tilde{\theta}$ via the Kramers-Moyal expansion, we actually only need to define the probability distribution of $\tilde{\Delta}(dt)$ when $t'-t=dt$ is infinitesimally small. We will show that we can (approximately) match the continuous-time update over time interval $\tau$ to the discrete-time update by defining the moments of $\tilde{\Delta}(dt)$ as:
\begin{gather}
  E[\tilde{\Delta}(dt)^\gamma] = \frac{\kappa_\gamma \big(\theta(t) \big)}{\tau}dt, \label{eq:alphaDef}
\end{gather}
with $\gamma \in \NN_{0}^p$ (where $\NN_{0}^p$ denotes $p$-dimensional non-negative integers). That is, the moments of the small-time increments of $\tilde{\theta}(t)$ are directly proportional to the cumulants of the discrete time process $\theta_j$ it seeks to approximate. (We will find that resulting process will approximate the discrete one well when $\kappa_\gamma \big(\theta(t) \big)$ does not change much in value within each $\tau$ time increment, so that the small time increments during that time period are approximately independent and identically distributed.)

Then we will show that (setting $\tau=1$ for simplicity now) the following Kramers-Moyal (KM) expansion describes the evolution of the distribution of $\tilde{\theta}$:
\begin{gather}
\partial_t \rho(\theta, t) = \sum_{\gamma \in \NN_1^p}  \frac{(-1)^\gamma}{\gamma!} \partial_\theta^\gamma \Big\{ \kappa_\gamma(\theta) \rho(\theta,t)\Big\}, \label{eq:km}
\end{gather}
where $\partial^\gamma_\theta f(\theta) = \partial^{\gamma_0}_{\theta_0} \ldots \partial^{\gamma_{p-1}}_{\theta_{p-1}} f(\theta)$ for an arbitrary function $f(\theta)$ (as is standard in multi-index notation), and $\NN_1^p$ denotes the $p$-dimensional positive integers.
The KM expansion can be truncated to second order to obtain a Fokker-Planck approximation of the discrete-time stochastic process.

\subsubsection*{Moments of the continuous time process in terms of cumulants of the discrete time process}

Our goal in this section is to justify Equation \ref{eq:alphaDef}. Let $dt = \tau/K$ for some positive integer $K$, and define $t_j=jdt$ for $j=0, \ldots, K,$ and $dt_j = t_{j+1} - t_j = dt.$ Recall that $\tilde{\Delta}(dt_j)$ is the small time increment of our continuous process between $t_{j-1}$ and $t_j,$ so that $\tilde{\Delta}(\tau, K) = \sum_{j=1}^K \tilde{\Delta}(dt_j)$ is the total change in our process over $\tau$ time units when breaking the temporal interval into $K$ equal-sized increments. We want to determine the moments of $\tilde{\Delta}(dt_j)$ so that $E[\tilde{\Delta}(\tau)^\gamma] = \lim_{K \rightarrow \infty} E[\tilde{\Delta}(\tau, K)^\gamma]$ matches the corresponding moment of the discrete SGD step $E[\Delta(\theta)]$.  Assuming the increments $\tilde{\Delta}(dt_j)$ are approximately i.i.d. within each time interval of length $\tau,$ we can approximate $\tilde{\Delta}(\tau, K)$ as a sum of i.i.d. random variables. With simplified notation, we model the problem as follows. Consider a sum $S_K = \sum_{i=1}^K X_i,$ where $X_i$ are i.i.d. random variables (each in $\RR^p$). Suppose that we know the `desired' limiting random variable $S,$ along with its cumulants. We want to find the moments of the i.i.d. random variables $X$ so that $\lim_{K \to \infty} S_K = S$. We will use here the following definitions. Let $m_\gamma^X$ and $\kappa_\gamma^X$ denote the moments and cumulants of an arbitrary random variable $X.$ Let $M_X(t) = E[e^{t^T X}]$ be the moment generating function of $X,$ and $C_X(t) = \log M_X(t)$ the cumulant generating function, so that $m_\gamma^X = \partial^\gamma M_X(t)|_{t=0}$ and $\kappa_\gamma^X = \partial^\gamma C_X(t)|_{t=0}$. We will also use the identity:
\begin{align}
    e^x = \lim_{K\to \infty} \big(1 + \frac{x}{K}\big)^K. \label{eq:exp_lim}
\end{align}
To restate the problem more precisely in this notation, we want to find $m_\gamma^X$ such that $\lim_{K \to \infty} C_{S_K}(t) = C_S(t)$. Using the i.i.d. assumption and identity \ref{eq:exp_lim}, we obtain:
\begin{align*}
    M_{S_K}(t) &= E[ e^{tS_K} ] = \prod_{i=1}^K E[ e^{t X_i} ] = (M_X)^K, \text{ so}\\
     \lim_{K \to \infty} M_{S_K}(t) &= \lim_{K \to \infty} (M_{X})^K = \lim_{K \to \infty} (1 + \xi/K)^K, \quad \text{where $\xi$ satisfies $M_X \equiv 1 + \xi/K$}\\
    &= e^{\xi}.
\end{align*}
So $\lim_{K \to \infty} C_{S_K}(t) = \xi,$
 and since we want $\lim_{K \to \infty} C_{S_K}(t) = C_S(t)$, we equate $\xi = C_S(t)$ to find that $M_X(t) = 1 + C_S(t)/K,$ so 
$$
    m_\gamma^X = \partial^\gamma M_X(t)|_{t=0} = \frac{1}{K} \partial^\gamma C_S(t)|_{t=0} = \frac{1}{K} \kappa_\gamma^S.
$$
In order to achieve the desired limit, we then need the moments of the i.i.d. variables $X$ to be equal to $1/K$ times the cumulants of the desired limiting distribution $S$. To explicitly connect this result back to the continuous approximation of SGD, we associate $X_i = \tilde \Delta(dt_i)$, $S_j = \tilde \Delta(\tau, K)$, $S = \Delta(\theta)$, and $dt = \tau/K$. Then $m_\gamma^X = \frac{1}{K} \kappa_\gamma^S$ translates to Equation \ref{eq:alphaDef}, as we desired. Note that the i.i.d. assumption on the increments $\tilde \Delta(dt_j)$ is an approximation that introduces error into (\ref{eq:alphaDef}). In the approximation, we assume that $P(\tilde \Delta(dt_j)) \approx P(\tilde \Delta(dt_0))$ for $t_j = j dt$, $j=1, \ldots, K$, $dt = \tau/K$. If $\tau$ is small and $\tilde \Delta$ varies slowly, then this condition will approximately hold.

\subsubsection*{The continuous Kramers-Moyal expansion}

Our goal in this section is to obtain the Kramers-Moyal expansion (\ref{eq:km}). The approach is essentially to Taylor-expand the Chapman-Kolmogorov equation, take the limit as $dt \to 0$, and use result (\ref{eq:alphaDef}) for the moments of the infinitesimal increments. The Markov assumption for $\tilde{\theta}(t)$ implies the continuous-time Chapman Kolmogorov equation 
\begin{gather}
\rho(\theta, t') = \int \rho(\theta - \Delta, t)W(\Delta, t'|\theta - \Delta, t) d \Delta, \label{eq:ck_continuous}
\end{gather}
where $W(\Delta, t'|\theta, t)= P \big(\tilde{\theta}(t') = \theta + \Delta | \tilde{\theta}(t) = \theta \big)$ is the transition probability function, and $t' \geq t.$ Substituting $t' = t + dt$, and using our definition of $\tilde \Delta$, we obtain
\begin{align}
\rho(\theta, t + dt) &= \int \rho(\theta - \Delta, t)W(\Delta, t+dt|\theta - \Delta, t) d \Delta, \label{eq:ck_dt}\\
\text{where } W(\Delta, t+dt|\theta, t) &= P \big(\tilde{\Delta}(dt) = \Delta | \tilde{\theta}(t) = \theta \big). \notag
\end{align}
Next we Taylor-expand the left-hand-side of (\ref{eq:ck_dt}) in $\Delta$ (recalling that in multi-index notation, the infinite Taylor expansion of $f(\theta)$ is
$f(\theta+h) = \sum_{\gamma \in \NN_0^p} \frac{\partial^\gamma_\theta f(\theta)}{\gamma!} h^\gamma$):
\begin{align}
\rho(\theta, t + dt) &= \int \sum_{\gamma \in \NN_0^p} \frac{(-\Delta)^\gamma}{\gamma!} \partial_\theta^\gamma \Big( \rho(\theta, t)W(\Delta, t+dt|\theta, t) \Big) d \Delta \notag\\
&= \sum_{\gamma \in \NN_0^p} \frac{(-1)^\gamma}{\gamma!} \partial_\theta^\gamma \Big( \rho(\theta, t) \int \Delta^\gamma W(\Delta, t+dt|\theta, t) d \Delta \Big) \notag\\
&= \sum_{\gamma \in \NN_0^p} \frac{(-1)^\gamma}{\gamma!}\partial_\theta^\gamma \Big( \rho(\theta, t) E[\tilde \Delta(dt)^\gamma] \Big) \notag\\
&= \rho(\theta, t) + \sum_{\gamma \in \NN_1^p} \frac{(-1)^\gamma}{\gamma!}\partial_\theta^\gamma \Big( \rho(\theta, t) E[\tilde \Delta(dt)^\gamma] \Big) \notag
\end{align}
Finally we take the limit as $dt \to 0$ and use (\ref{eq:alphaDef}):
\begin{align}
\implies \partial_t \rho(\theta, t) &= \lim_{dt \to 0} \frac{\rho(\theta + \Delta, t + dt) - \rho(\theta + \Delta, t)}{dt} \notag\\
&= \sum_{\gamma \in \NN_1^p} \frac{(-1)^\gamma}{\gamma!}\partial_\theta^\gamma \Big( \rho(\theta, t) \lim_{dt \to 0} \frac{E[\tilde \Delta(dt)^\gamma]}{dt}  \Big) \notag\\
&= \sum_{\gamma \in \NN_1^p}  \frac{(-1)^\gamma}{\tau \gamma!} \partial_\theta^\gamma \Big\{ \kappa_\gamma(\theta) \rho(\theta,t)\Big\}, \notag
\end{align}
The result above is Equation \ref{eq:km}.

\subsubsection*{The continuous process approximates the discrete process}
Now that we have defined $\tilde \theta$ via its Kramers-Moyal expansion (\ref{eq:km}), we need to check that it approximates SGD in the sense of Equation \ref{eq:approxP} (we need to show that
$\rho(\theta, j\tau) \approx P(\theta_j=\theta)$).
We seek to understand the approximation error in terms of the difference between $ \tilde{m}_\gamma(\theta)$ and $m_\gamma(\theta).$ The error arises because $ \tilde{m}_\gamma(\theta)$ relies on the approximation that small-time increments of $\tilde{\theta}(t)$ are i.i.d.. We assume that $\rho(\theta, t) = P(\theta_j=\theta),$ and then study $\rho(\theta, t+\tau),$ starting from the Chapman-Kolmogorov for $\tilde{\theta}(t)$ (\ref{eq:ck_continuous}):
\begin{align}
\rho(\theta, t+\tau) &= \int \rho(\theta - \Delta, t)W(\Delta, t+\tau|\theta - \Delta, t) d \Delta \nonumber \\ 
&= \int P(\theta_j=\theta - \Delta)\biggr( P(\Delta_j = \Delta | \theta_j = \theta - \Delta) \nonumber \\
& \quad +
\big\{W(\Delta, t+\tau|\theta - \Delta, t)  - P(\Delta_j = \Delta | \theta_j = \theta - \Delta)\big\}\biggr) d \Delta \nonumber \\ 
&= P(\theta_{j+1}=\theta) + \int P(\theta_j=\theta - \Delta)\big\{W(\Delta, t+\tau|\theta - \Delta, t)  - P(\Delta_j = \Delta | \theta_j = \theta - \Delta)\big\}d \Delta  \nonumber \\ 
&= P(\theta_{j+1}=\theta) + \int \rho(\theta - \Delta, t) W(\Delta, t+\tau|\theta - \Delta, t)d \Delta  \nonumber \\
& \quad - \int P(\theta_j=\theta - \Delta) P(\Delta_j = \Delta | \theta_j = \theta - \Delta)d \Delta.
\label{eq:ck_continuous2}
\end{align}
The two integrals above can be Taylor-expanded in the same way as in our derivation of the Kramers-Moyal expansion to yield
\begin{align}
 e(\theta) &= \rho(\theta, t+\tau) - P(\theta_{j+1}=\theta) 
 = \sum_{\gamma \in \NN_0^p} \frac{(-1)^\gamma}{\gamma!}\partial_\theta^\gamma \big\{\rho(\theta, t)E[\tilde{\Delta}(\tau)^\gamma] - P(\theta_{j}=\theta)E[\Delta_j^\gamma]\big\} \nonumber \\
 &= \sum_{\gamma \in \NN_0^p}\frac{(-1)^\gamma}{\gamma!} \partial_\theta^\gamma \big\{\rho(\theta, t)\big( \tilde{m}_\gamma(\theta) - m_\gamma(\theta) \big)\big\} = \sum_{\gamma \in \NN_0^p} \frac{(-1)^\gamma}{\gamma!} \partial_\theta^\gamma \big\{\rho(\theta, t) e_\gamma(\theta)\big\},
\label{eq:thetas_approx_error}
\end{align}
where we define the error in the $\gamma$-th moment to be $e_\gamma(\theta)= \tilde{m}_\gamma(\theta) - m_\gamma(\theta).$ So, i.e.,
\begin{align}
 \int |e(\theta)|d \theta & \leq \sum_{\gamma \in \NN_1^p} \frac{1}{\gamma!} |\partial_\theta^\gamma \big\{\rho(\theta, t) e_\gamma(\theta)\big\}|.
\label{eq:thetas_abs_net_error}
\end{align}
Recall that the errors $e_\gamma(\theta)$ are due to the i.i.d. assumption on the infinitesimal increments in the continuous approximation; that is, the assumption that the distribution of $\tilde \Delta$ is approximately constant in an interval of length $\tau$. Therefore, if $\tilde \Delta$ varies slowly relatively to the timescale $\tau$, the approximation will be close.

\subsection{Continuous-Time Diffusion Approximation of SGD}
We now choose $\Delta$ to be as in Equation \ref{eq:SGD} to study SGD. Our goal here is to approximate SGD with a continuous-time process. Because cumulants of independent random variables are additive, and letting $\omega_\gamma(\theta)$ be the $\gamma$-th cumulant of $\partial_\theta U(x_i, \theta)$ (i.e., the gradient evaluated at a single sample of $x$),  we find that the cumulant for the minibatch of size $B$ is
  $$\kappa_\omega(\theta) = \big(-T\big)^{|\gamma|} B \omega_\gamma(\theta).$$ 
So the KM expansion (\ref{eq:km}) of the continuous-time SGD approximation then becomes:
  \begin{align}
\partial_t \rho(\theta, t) &= B \sum_{\gamma \in \NN_1^p} \frac{T^{|\gamma|}}{\gamma!}\partial_\theta^\gamma \Big( \omega_\gamma(\theta) \rho(\theta, t) \Big). \label{eq:km_continuous2}
\end{align}
Clearly, then, as $T$ gets small, fewer terms in the expansion matter. The first two cumulants for SGD are the mean and variance of the gradients of $U$ over the training distribution, respectively:
\begin{align*}
    \omega_1(\theta) &= E_{x \sim \textit{tr}} [\partial_\theta U(x, \theta)], \quad
    \omega_2(\theta) = \text{Cov}_{x \sim \textit{tr}} [ \partial_\theta U(x, \theta)].
\end{align*}

When $T$ is small enough, or when the cumulants $\omega_\gamma(\theta)$ are small for $|\gamma|> 2$ (i.e., when $\partial_\theta U(x_i, \theta)$ is Gaussian, cumulants higher than 2 are zero), we can approximate the KM expansion of (\ref{eq:km_continuous2}) by the Fokker-Planck (FP) equation that retains only the first two terms in the expansion. Switching to the notation of the main text:
$$ \omega_1(\theta) = \partial_\theta U^\textit{tr}(\theta), \quad \omega_2(\theta) = D^\textit{tr}(\theta), $$
(so $D^\textit{tr}(\theta)=\omega_2(\theta) \in \mathcal{R}^{p \times p}$ is now the empirical covariance matrix of the gradients in the training, or \textit{diffusion} matrix, and $U^\textit{tr}(\theta)$ is the train loss), we obtain
\begin{align}
\partial_t \rho(\theta, t) 
&:= \lambda \sum_{i=1}^p \partial_{\theta_i} \Big\{ \partial_{\theta_i} U^\textit{tr}(\theta) \rho(\theta,t)\Big\} + \frac{1}{2} \lambda T \sum_{i, j=1}^p \partial^2_{\theta_i, \theta_j} \Big\{ D^\textit{tr}_{ij}(\theta) \rho(\theta,t)\Big\}. \nonumber
\end{align}
Rewriting using the divergence operator $\partial_\theta \cdot v = \sum_{i=1}^p \partial_{\theta_i} v(\theta)$, we obtain the Fokker-Planck SGD equation stated in the main text (\ref{eq:fp}).

\subsubsection*{When is the SGD diffusion approximation accurate?}
The FP approximation of SGD follows from two approximations. First, we need the increments $\tilde \Delta(dt_j)$ within a time interval of $\tau$ (corresponding to a single SGD update) to be approximately i.i.d., so that the KM expansion (\ref{eq:km}) is an accurate approximation of the discrete time SGD process. This condition holds when the product of the expected change in $\theta$ during a single update and of the derivative w.r.t. $\theta$ of the density of any small increment is small.\footnote{
To see this, consider the joint probability of two infinitesimal increments within the same time period of length $\tau,$ i.e.,
$P\big(\tilde \Delta(dt_0) =x_0, \tilde \Delta(dt_j) = x_j \big)=P\big(\tilde \Delta(dt_0) =x_0\big) P\big( \tilde \Delta(dt_j) =x_j | \tilde \Delta(dt_0) =x_0\big),$
where $j > 0,$ and assume $\theta=\theta_0$ at the beginning of the $dt_0$ interval. We can write the latter probability as
\begin{align}
    P\big( \tilde \Delta(dt_j) = x_j | \tilde \Delta(dt_0) = x_0 \big) =E_{W}[P\big( \tilde \Delta(dt_j) = x_j | \theta_{j-1}=\theta_0+W\big)], \text{ with } W = x_0 + \tilde \Delta(t_1, t_{k-1}) \nonumber
\end{align}
is a random variable describing the increment since the start of the $\tau$ time interval until the beginning of the $k$-th infinitesimal increment $t_{k-1}.$ Letting $\mu_w$ denote the mean of $W,$ we can Taylor expand to first order around $W=0$ to find that
\begin{align*}
    E_{W}[P\big( \tilde \Delta(dt_j) = x_j | \theta_{j-1}=\theta_0+W\big)] \approx P\big( \tilde \Delta(dt_j) = x_j | \theta_{j-1}=\theta_0\big) + \mu_w' \partial_{\theta_{j-1}} P\big( \tilde \Delta(dt_j) = x_j | \theta_{j-1}=\theta_0\big)
\end{align*}
plus higher order terms. When the second term above is small compared to the first, we can finally write that $P\big(\tilde \Delta(dt_0) = x_0, \tilde \Delta(dt_j) = x_j \big)=P\big(\tilde \Delta(dt_0) = x_0 \big) P\big( \tilde \Delta(dt_j) = x_j \big),$ and since $|dt_j|=|dt_0|=dt$, and these increments only depend on these magnitudes and on $\theta$ at the beginning of each increment (both equal to $\theta_0$ in our expansion above), then the i.i.d. assumption is satisfied. Conversely, when $\mu_w$ and/or $ \partial_\theta P\big( \tilde \Delta(dt_j) = x_j | \theta=\theta_0\big)$ are large we expect the increments not to be i.i.d., and the KM expansion to not approximate the discrete process well.} The second condition required for the FP equation to hold is that terms of order $|\gamma|>2$ in the KM expansion must be small (relative to the $|\gamma| \leq 2$ terms), so that the truncation that yields the FP equation is appropriate. The latter is satisfied when $T$ is small, and/or the third and higher cumulants of the gradients  (i.e., $\omega_\gamma(\theta)$ for $|\gamma|>2$) are small. Finally, our steady-state analysis in the next section relies on the distribution having actually reached steady-state, and the number of SGD steps required to reach steady-state scales inversely with temperature, so in practice with very small temperature the steady-state could be difficult to attain. Therefore, practically speaking, we expect the SGD diffusion approximation to start to break down for large $T$ (i.e.\ large ratio of LR to batch size), or for fixed small $T$ at extreme (very small or very large) learning rates or batch sizes. A large learning rate makes the expected change in $\theta$ in a single update large, potentially violating the i.i.d. assumption of the small time increments, while a large batch size at small fixed temperature also implies a large learning rate, and has the same effect. We similarly expect the mean change in an SGD update to be large at the beginning of an SGD run, and our FP approximation to not be valid during some initial transient period. Lastly, the i.i.d. assumption can be violated when the derivative of the mean and covariance of a single SGD update with respect to $\theta$ is large.

\section{Steady State of the SGD Diffusion Approximation}

\subsection{Steady State Distribution}
\label{append:steady-state}

Our goal in this section is to show that (\ref{eq:sgdss}) is the steady-state solution of our SGD diffusion approximation (\ref{eq:fp}), and clarify the assumptions under which this is true. Our solution of the Fokker-Planck equation is identical to one presented in \cite{gardiner2009} Section 6.2. We consider the underparametrized case (Section \ref{sec:overparam} extends to the overparametrized case), so that $D^\textit{tr}(\theta)$ is invertible and the SGD drift term spans $\RR^p$ even without regularization. We assume the process is ergodic, and therefore has a single steady-state distribution $\rho(\theta).$ To find it, we first set (\ref{eq:fp}) to zero and obtain the steady-state condition $\partial_\theta \cdot J(\theta)=0$. A distribution $\rho(\theta)$ where $J$ is constant, if it can be found, satisfies the above. We also now require that $\rho(\theta) \to 0$ and $\partial_\theta \rho(\theta) \to 0$ as $\theta \to \pm \infty$, so we attempt to find a solution that satisfies $J(\theta) = 0$ everywhere, and where $J(\theta)= \partial_{\theta} U^\textit{tr}(\theta) \rho(\theta) + \frac{T}{2} \partial_\theta \cdot \big( D^\textit{tr}(\theta) \rho(\theta) \big).$ Some algebra then yields the steady-state solution
\begin{align}
    \frac{\partial_\theta \rho(\theta, t)}{\rho(\theta, t)} &= - \frac{2}{T} D^\textit{tr}(\theta)^{-1} \big(\partial_{\theta} U^\textit{tr}(\theta) + \frac{T}{2}\partial_\theta \cdot D^\textit{tr}(\theta) \big) := -\frac{2}{T} \mathcal{V}(\theta), \text{ so} \nonumber \\
    \int^\theta \partial_\theta \log \rho(\xi) \cdot d\xi &= -\frac{2}{T} \int^\theta \mathcal{V}(\xi) \cdot d\xi := -\frac{2}{T} v(\theta, T), \text{ which can be re-arranged as} \nonumber \\
    \rho(\theta) \propto  e^{-\frac{2}{T} v(\theta, T)},& \text{ with } v(\theta, T) = \int^\theta D^\textit{tr}(\xi)^{-1} \big( \partial_{\xi} U^\textit{tr}(\xi) + \frac{T}{2}\partial_\theta \cdot D^\textit{tr}(\xi) \big) \cdot d\xi. \label{eq:rho_x}
\end{align}
This solution relies on a line integral to define $v(\theta, T)$ that is path independent, and therefore needs the following assumption
\begin{assumption}
\label{assump:curl}
$\mathcal{V}(\theta) = D^\textit{tr}(\theta)^{-1} \big( \partial_{\theta} U^\textit{tr}(\theta) + \frac{T}{2}\partial_\theta \cdot D^\textit{tr}(\theta) \big)$ is a gradient; that is the curl of $\mathcal{V}$ vanishes.
\end{assumption}
In other words, since $\mathcal{V}$ is defined as $-\partial_\theta \log \rho(\theta)$, the first equation above can only be satisfied if $\mathcal{V}$ is a gradient; a necessary and sufficient condition for this is the vanishing of the curl or so-called potential conditions
$\partial_{\theta_j} \mathcal{V}_i = \partial_{\theta_i} \mathcal{V}_j$ where $\mathcal{V}_i$ denotes the $i$-th entry of $\mathcal{V}$. When these conditions hold, the Hessian of $\log \rho(\theta)$ is symmetric, as it should be. So we assume that $\partial_{\theta} U^\textit{tr}(\theta), D^\textit{tr}(\theta)$ and are such that the assumption above holds.

\subsubsection*{Examples where Assumption \ref{assump:curl} holds}
Assumption \ref{assump:curl} clearly holds in the case where $D^\textit{tr}$ is constant and isotropic (i.e. $D^\textit{tr} = c^{-1}I$ for positive constant $c$), since then $\mathcal{V}(\theta) = c \partial_{\theta} U^\textit{tr}(\theta)$, which is the gradient of $c U^\textit{tr}$. But it can also hold for nonconstant, anisotropic $D^\textit{tr}$ as well, as the following examples confirm. 
\begin{enumerate}
    \item 
Isotropic but non-constant noise: consider  $D^\textit{tr}(\theta) = U^\textit{tr}(\theta) I,$  so that $\partial_\theta \cdot D^\textit{tr}(\theta) = \partial_\theta U^\textit{tr}(\theta)$. Then $\mathcal{V}(\theta)=(1+\frac{T}{2})\partial_\theta \log U(\theta)$ is a gradient, and we have
$v(\theta, T) = (1 + \frac{T}{2})\log(U(\theta))$, and $\rho(\theta) \propto U(\theta)^{-(1+\frac{2}{T})}.$ This argument works more generally: let $f: \RR \to \RR$ be any smooth scalar function, and let $D^\textit{tr}(\theta) = \frac{1}{f'(U(\theta))} I$ where $f'(U(\theta))=\partial_{U(\theta)} f(U(\theta))$ (the first example is a special case with $f = \log U$). Then the chain rule implies that $(D^\textit{tr}(\theta))^{-1}  \partial_\theta U^\textit{tr}(\theta) = \partial_\theta f(U(\theta)).$ Similarly, $(D^\textit{tr}(\theta))^{-1}(\partial_\theta \cdot D^\textit{tr}(\theta)) = -\partial_\theta \log f'(U(\theta)),$ so $\mathcal{V}= \partial_\theta f(U(\theta)) - \frac{T}{2}\partial_\theta \log  f'(U(\theta)),$ $v(\theta, T)= f(U(\theta)) - \frac{T}{2}\log f'(U(\theta)),$ and $\rho(\theta) \propto e^{-\frac{2}{T}f(U(\theta)) + \log f'(U(\theta))}.$  
\item Anisotropic but constant noise: Consider $D^\textit{tr}(\theta) = \partial_\theta^2 U^\textit{tr}(\theta)^{-1}$ and suppose that $U^\textit{tr}(\theta)$ is quadratic, so that $D^\textit{tr}$ is constant and $\partial_\theta \cdot D^\textit{tr} = 0$. Then \\
$\partial_\theta \big( \half \partial_\theta U^\textit{tr}(\theta)^T \partial_\theta U^\textit{tr}(\theta) \big) = \partial_\theta^2 U^\textit{tr}(\theta) \partial_\theta U^\textit{tr}(\theta) = (D^\textit{tr})^{-1} \partial_\theta U^\textit{tr}(\theta) = \mathcal{V}(\theta),$
which shows that $\mathcal{V}$ is a gradient. In this case, $v(\theta) = \half \partial_\theta U^\textit{tr}(\theta)^T \partial_\theta U^\textit{tr}(\theta)$.
\item Non-constant and anisotropic noise: Let $U^\textit{tr}(\theta) = \sum_i U_i(\theta_i)$ and suppose that $D^\textit{tr}(\theta) = \text{diag}\{ d_i(\theta_i) \}$, where $d_i(\theta_i) = \frac{1}{f'(U_i(\theta_i)}$ for an arbitrary smooth scalar function $f.$ Note that both $U_i$ and $d_i$ are functions only of $\theta_i$. Then $\big( (D^\textit{tr})^{-1} \partial_\theta U^\textit{tr} \big)_i  = f'(U_i) \partial_{\theta_i} U_i = \partial_{\theta_i} f(U_i)$. Also, $\big( (D^\textit{tr})^{-1} \partial_\theta \cdot D^\textit{tr} \big)_i = \frac{\partial_{\theta_i} d_i}{d_i(\theta_i)} = \partial_{\theta_i} \log d_i(\theta_i) = -\partial_{\theta_i} \log f'(U_i(\theta_i)$. So $\mathcal{V}_i = \partial_{\theta_i} f(U_i) - \frac{T}{2} \partial_{\theta_i} \log f'(U_i(\theta_i)$. Defining $v(\theta) = \sum_i f( U_i(\theta_i)) - \frac{T}{2}\sum_i \log f'(U_i(\theta_i))$, we get $\partial_\theta v(\theta) = \mathcal{V}(\theta)$. This example results in independent dynamics for each parameter, which is unlikely to hold for realistic models, but it does show that Assumption 1 holds for more models with just those with constant isotropic noise. (As an additional example, note that in example 2 where $D^\textit{tr}(\theta) = \partial_\theta^2 U^\textit{tr}(\theta)^{-1}$, we obtain the same $\mathcal{V}$ if we remove quadratic $U^\textit{tr}$/constant $D^\textit{tr}$ assumption but take the limit as $T \to 0$ to drop the $\partial_\theta \cdot D^\textit{tr}$ term.)
\end{enumerate}
Further work is necessary to characterize all the situations for which SGD satisfies Assumption \ref{assump:curl}. \cite{chaudhari2018stochastic} state in effect that $J \neq 0$ whenever $D^\textit{tr}$ is nonconstant and/or anisotropic, so the examples above are counterexamples since they all satisfy $J=0$ (a consequence of Assumption \ref{assump:curl}).

\subsection{Gaussian Mixture Approximation}
\label{append:sgd_ss_gmm}

We further approximate the steady-state distribution of SGD by a mixture of Gaussians, i.e., as a distribution of the form $\rho(\theta)=\sum_k w_k \mathcal{N}\big(\mu_k, \Sigma_k \big).$ So we need to find the means $\mu_k$, covariances $\Sigma_k$, and weights $w_k$. In our derivation, we assume the underparametrized case, where for example the gradient covariance and Hessian are full rank. However, the arguments are similar in the overparametrized case using the modified SGD (\ref{eq:SGDMod}).
From (\ref{eq:sgdss}), the first two derivatives of $v$ are
\begin{align*}
\partial_{\theta} v({\theta}, T) &= D^\textit{tr}(\theta)^{-1} \big( \partial_{\theta} U^\textit{tr}(\theta) + \frac{T}{2}\partial_{\theta} \cdot D^\textit{tr}(\theta) \big) \\
\partial_{\theta}^2 v({\theta}, T) &= \partial_{\theta} (D^\textit{tr}(\theta)^{-1}) \big( \partial_{\theta} U^\textit{tr}(\theta) + \frac{T}{2}\partial_{\theta} \cdot D^\textit{tr}(\theta) \big) 
+ D^\textit{tr}(\theta)^{-1} \big( \partial_{\theta}^2 U^\textit{tr}(\theta) + \frac{T}{2} \partial_{\theta}^T (\partial_{\theta} \cdot D^\textit{tr}(\theta)) \big)
\end{align*}
If $\mu$ is any local extremum of $v$, then
\begin{align*}
0 = \partial_{\theta} v(\mu, T) &= D^\textit{tr}(\mu)^{-1} \big( \partial_{\theta} U^\textit{tr}(\mu) + \frac{T}{2}\partial_{\theta} \cdot D^\textit{tr}(\mu) \big) \\
\partial_{\theta}^2 v(\mu, T) &= D^\textit{tr}(\mu)^{-1} \big( \partial_{\theta}^2 U^\textit{tr}(\mu) + \frac{T}{2} \partial_{\theta}^T (\partial_{\theta} \cdot D^\textit{tr}(\mu)) \big),
\end{align*}
(where the first term of $\partial_{\theta}^2 v({\theta}, T)$ drops out since $D^\textit{tr}$ is positive definite, so $\partial_{\theta} v(\mu, T) = 0 \implies \partial_{\theta} U^\textit{tr}(\mu) + \frac{T}{2}\partial_{\theta} \cdot D^\textit{tr}(\mu) = 0$). Now, let ${\theta}^\textit{tr}_k$ be a local minimum of $U^\textit{tr}$, and $\mu_k$ a nearby local minimum of $v$. To find the approximate bias $b_k = {\theta}^\textit{tr}_k - \mu_k$, we can expand the $0 = \partial_{\theta} v(\mu, T)$ equation above:
\begin{align*}
    0 =& D^\textit{tr}(\mu_k)^{-1} \big( \partial_{\theta} U^\textit{tr}(\mu_k) + \frac{T}{2}\partial_{\theta} \cdot D^\textit{tr}(\mu_k) \big)
    \approx D^\textit{tr}(\mu_k)^{-1} \big( -C_k^\textit{tr} b_k + \frac{T}{2}\partial_{\theta} \cdot D^\textit{tr}(\mu_k) \big),
\end{align*}
since $\partial_{\theta} U^\textit{tr}(\mu_k) \approx \partial_{\theta} U^\textit{tr}({\theta}^\textit{tr}_k) + \partial_{\theta}^2 U^\textit{tr}(\theta^\textit{tr}_k) (\mu_k - {\theta}^\textit{tr}_k) = -C_k b_k.$ So 
$$b_k \approx \frac{T}{2} (C_k^\textit{tr})^{-1} (\partial_{\theta} \cdot D_{\theta}(\mu_k)) = O(T).$$
For the covariance, we can make a Laplace approximation to $\rho(\theta)$ centered at $\mu_k$:
\begin{align}
v({\theta}, T) &\approx v(\mu_k, T) + \half \partial_{\theta}^2 v(\mu_k, T) , \text{ so}\nonumber\\
\rho(\theta) &\propto e^{-\frac{2}{T} v({\theta}, T)}
\approx e^{-\frac{2}{T} v(\mu_k, T)} e^{-\frac{1}{T} \partial_{\theta}^2 v(\mu_k, T)} \nonumber\\
&\propto \sqrt{|\Sigma_k|} \mathcal{N}( \mu_k, \Sigma_k), \text{ with } \Sigma_k = \frac{T}{2} (\partial_{\theta}^2 v(\mu_k, T))^{-1}.
\end{align}
Therefore, $\Sigma_k$ is also $O(T)$ (and inversely proportional to the curvature of the effective potential). The Laplace approximation also allows us to find the weights. We use the eigendecomposition of $\partial_{\theta}^2 v(\mu_k, T)$ to approximate the integral of $\rho(\theta)$ over the basin $\mathcal{B}_k$:
\begin{align*}
    \partial_{\theta}^2 v(\mu_k, T) &= M_k \Lambda_k M_K^T \quad \text{($\partial_{\theta}^2 v(\mu_k, T)$ is positive definite so $\lambda_i > 0$ for all $i$)} \\
    q = M_k^T ({\theta} - \mu_k) &\implies 
    e^{-\frac{1}{T} ({\theta} - \mu_k)^T \partial_{\theta}^2 v(\mu_k, T) ({\theta} - \mu_k) } = e^{-\frac{1}{T} q^T \Lambda_k q } \\
    \int_{-\infty}^\infty e^{-\frac{1}{T} q^T \Lambda_k q } dq 
    &= \prod_{i=1}^n \int_{-\infty}^\infty e^{-\frac{1}{T} \lambda_i q_i^2 } dq_i 
    = \prod_{i=1}^n \sqrt{\pi / \lambda_i} = \sqrt{\pi^n |\Lambda_k|^{-1}} 
    = \sqrt{\pi^n |\partial_{\theta}^2 v(\mu_k, T)|^{-1}}\\
    \implies w_k &= \int_{\mathcal{B}_k} \rho(\theta) d\theta \approx 
    e^{-\frac{2}{T}v(\mu_k)} \int_{\mathcal{B}_k} e^{-\frac{1}{T} ({\theta} - \mu_k)^T \partial_{\theta}^2 v(\mu_k) ({\theta} - \mu_k) } d\theta \\
    &= e^{-\frac{2}{T}v(\mu_k)} \int_{-\infty}^\infty e^{-\frac{1}{T} q^T \Lambda_k q } dq
    = e^{-\frac{2}{T}v(\mu_k)} \sqrt{\pi^n |\partial_{\theta}^2 v(\mu_k, T)|^{-1}} \\
    &\propto e^{-\frac{2}{T}v(\mu_k)} \cdot |\partial_{\theta}^2 v(\mu_k, T)|^{-\half}.
\end{align*}
This is Equation \ref{eq:weight_rho} in the main text. It tells us that the basin weights depends on the depth and width of the basin in terms of the \emph{effective potential}.

\section{Reparametrization Invariance}
\label{append:reparam}

\subsection*{Expected test loss under any parameter distribution is reparametrization-invariant}

Let $U(\theta)$ denote any loss function (for example, the test loss, or the train loss with or without regularization), and let $\rho(\theta)$ be any distribution over the model parameters. Consider a reparametrization $y = r^{-1}(\theta)$ of $\theta$ (with $\theta, y \in \RR^p$), where $r:\RR^{p} \to \RR^p$ is invertible.  For an arbitrary function $f(\theta)$, we define the reparametrized version by $f^r(y) = f(r(y)) = f(\theta)$. With this notation, we want to show that $E_{\theta \sim \rho}[U(\theta)]$ is reparametrization-invariant.

Let $\rho_Y(y)$ denote the p.d.f. of $y = r^{-1}(\theta)$, and note that (using a general formula for invertible functions of random variables)
$\theta = r(y) \implies \rho_Y(y) = \rho(\theta) \big| \frac{dr}{dy} \big|$. Then:
\begin{align*}
    E_{\theta \sim \rho}[U(\theta)] &= \int_{-\infty}^\infty  U(\theta) \rho(\theta) d\theta \\
    &= \int_{-\infty}^\infty  U(r(y)) \rho(r(y)) \biggr| \frac{dr}{dy} \biggr| dy, \quad \text{(integral change of variable $y = r^{-1}(\theta)$)} \\ 
    &= \int_{-\infty}^\infty U^{r}(y) \rho_Y(y) dy, \quad \text{using $\rho_Y(y) = \rho(\theta)\biggr| \frac{dr}{dy} \biggr|$} \\
    &= E_{y \sim \rho_Y}[U^{r}(y)]
\end{align*}
Therefore $E_{\theta \sim \rho}[U(\theta)]$ is reparametrization-invariant.

In particular, this means that $E_{\theta \sim \rho}[U^\textit{test}(\theta)]$ is reparametrization-invariant, where $U^\textit{test}(\theta)$ is the test loss and $\rho$ is the SGD steady-state distribution.

\subsection*{Taylor approximation is reparametrization-invariant}

Using Equations \ref{eq:taylor_test_loss} and \ref{eq:approxAvgTestLoss}, we obtain the following approximation for the expected test loss:
\begin{align}
    E_{\theta \sim \rho}[U^{\textit{test}}(\theta)] \approx \sum_k w_k (U^{\textit{test}}_k + \half (s_k + b_k)^T C^{\textit{test}}_k (s_k + b_k) + \half \text{Tr} [\Sigma_k C^{\textit{test}}_k]) \label{eq:general_taylor_gm}
\end{align}

Our goal in this section is to show that this expression is approximately reparametrization-invariant for any distribution $\rho(\theta) = \sum_k w_k \mathcal{N}\big(\mu_k, \Sigma_k\big)$ such that $b_k$ and $s_k$ are both small (which holds for SGD because we are already assuming that $s_k$ is small, and the SGD steady-state distribution satisfies $b_k = O(T) \ll 1$). As before, let $r:\RR^p \to \RR^p$ be an invertible reparametrization. Defining $y_k^\textit{test} = r^{-1}(\theta^\textit{test}_k)$, we need to show that the terms appearing on the right-hand-side of (\ref{eq:general_taylor_gm}) are (approximately) reparametrization-invariant.

We can show that the weights $w_k$ are reparametrization-invariant using a similar argument to $E_{\theta \sim \rho}[U(\theta)]$:
\begin{align*}
w_k &= \int_{\mathcal{B}_k} \rho(\theta) d\theta 
= \int_{r^{-1}(\mathcal{B}_k)} \rho(r(y)) \biggr| \frac{dr}{dy} \biggr| dy 
= \int_{\mathcal{B}_k^y} \rho_Y(y) dy = w_k^r.
\end{align*}
Now we want to show that the remaining terms in left hand side of (\ref{eq:general_taylor_gm}) are approximately reparametrization-invariant. First, since the test loss is a scalar function, it is equal to its reparametrization, i.e.:
$$U_k^{\textit{test}, r}(y_k^\textit{test}) = U^{\textit{test}}(r(y_k^\textit{test})) \equiv U^{\textit{test}}(\theta^\textit{test}_k) = U_k^{\textit{test}}.$$
So $w_k U_k^{\textit{test}}$ is invariant. To help with the other terms, note that the first two derivatives of any scalar function $f: \RR^n \to \RR$ are:
\begin{align*}
\partial_y f(r(y)) &=  \partial_y r(y)^T \partial_\theta f(\theta),   \\
\partial_y^2 f(r(y)) 
&= \partial_y r(y)^T \partial_\theta^2 f(\theta) \partial_y r(y) \text{, if $\theta$ is a local extremum of $f$,}
\end{align*}
where $\partial_y r(y) = [\frac{\partial \theta_i}{\partial y_j}]_{i,j} \in \RR^{n \times n}$,  $\partial_y f(r(y)) \in \RR^{n}$, $\partial_y^2 f(r(y)) \in \RR^{n \times n}$, $\partial_\theta f \in \RR^{n}$, $\partial_\theta^2 f \in \RR^{n \times n}$.
We use these to find:
\begin{align*}
    C_k^{\textit{test},r} &= \partial_y^2 U^{\textit{test}}(r(y_k^\textit{test})) \nonumber\\
    &= \partial_y r(y_k^\textit{test})^T \partial_\theta^2 U^{\textit{test}}(\theta^\textit{test}_k) \partial_y r(y_k^\textit{test}) \quad \text{since $\partial_\theta U^{\textit{test}}(\theta^\textit{test}_k) = 0$} \nonumber\\
    &= \partial_y r(y_k^\textit{test})^T C_k^{\textit{test}} \partial_y r(y_k^\textit{test}) \\
    s_k^r &= y_k^\textit{tr} - y_k^\textit{test} \nonumber\\
    & \approx \partial_y r(y_k^\textit{test})^{-1}(\tilde \theta^\textit{test}_k - \theta^\textit{test}_k) + O((y_k^\textit{tr} - y_k^\textit{test})^2) \quad \text{since } \Delta \theta \approx (\partial_y r) \Delta y + O(\Delta y^2) \nonumber\\
    &= \partial_y r(y_k^\textit{test})^{-1} s_k + O(s_k^2) \\
    \Sigma_k^r &= \partial_y r(y_k^\textit{test})^{-1} \Sigma_k \partial_y r(y_k^\textit{test})^{-T} \quad \text{since } \rho_Y(y) = d_y r(y) \rho(\theta) \\
    b_k^r &= r^{-1}(\mu_k) - y^\textit{tr}_k \\
    &\approx \partial_y r(y_k^\textit{test})^{-1} b_k
\end{align*}
\begin{align}
    \text{Tr}\big(C_k^{\textit{test},r} \Sigma_k^r  \big) &= \text{Tr}\big(\partial_y r(y_k^\textit{test})^T C_k^{\textit{test}} \partial_y r(y_k^\textit{test}) \partial_y r(y_k^\textit{test})^{-1} \Sigma_k^{-1} \partial_y r(y_k^\textit{test})^{-T} \big) \nonumber\\
    &= \text{Tr}\big(C_k^{\textit{test}} \Sigma_k^{-1} \big) \nonumber\\
    s_k^{r^T} C_k^{\textit{test},r} s_k^r &= s_k^T \partial_y r(y)^{-T} \partial_y r(y_k^\textit{test})^T C_k^{\textit{test}} \partial_y r(y_k^\textit{test}) \partial_y r(y)^{-1} s_k + O(s_k^2) \nonumber\\
    &= s_k^T C_k^{\textit{test}} s_k + O(s_k^2) \nonumber\\
    b_k^{rT} C_k^{\textit{test},r} b_k^r &\approx b_k^T C_k^{\textit{test}} b_k, \quad b_k^{r^T} C_k^{test, r} s_k^r \approx b_k^T C_k^{\textit{test}} s_k \quad \text{similarly}
\end{align}
Since we have seen that each of the terms appearing in (\ref{eq:general_taylor_gm}) (i.e. $w_k$, $U^{\textit{test}}_k$, etc.) are all reparametrization-invariant, we conclude the whole expression for the expected test loss is reparametrization-invariant.

\section{The Overparametrized Case}
\label{append:overparametrization}

\subsection{Some Intuition}
\label{append:overparam_intuit}
One of the important properties of the overparametrized case (more model parameters than training samples) is that the loss landscape necessarily has zero- or near-zero-curvature directions, since in the absence of $\ell_2$-regularization there can be no more curved directions than there are training samples, and $\ell_2$-regularization adds a small amount of curvature in all directions. A question, then, is how zero-curvature directions affect the conclusions in this paper. As a rough intuition, we argue that zero-curvature directions may have little impact on test performance, and be deterministically forced to zero by SGD at steady state in the presence of any $\ell_2$ regularization, making them essentially irrelevant in the context of our analysis. Consider a simplified case where the set of nonzero-curvature directions is constant (rather than depending on $\theta$, as it does in general) and the same for train and test. Suppose that the train loss has a small amount of $\ell_2$-regularization (as is typical in practice) but the test loss does not. In this case, the train local minima are strict (single points) but the the test local minima are subspaces that include all the zero-curvature directions. The Taylor approximation (\ref{eq:taylor_test_loss}) is identical for \emph{any} point $\theta_k^\textit{test}$ in the test local minimum space spaces; that is, only the nonzero-curvature directions impact test performance or its shift-curvature approximation. Furthermore, SGD converges to a solution where the parameters in the zero-curvature directions are all equal to zero (as there is no diffusion in these directions and the drift term is $-\alpha \theta$, where $\alpha > 0$ is the $\ell_2$ parameter). Therefore, the zero-curvature directions are essentially irrelevant in the context of generalization and SGD.

Of course, the situation becomes more complex when we allow the nonzero-curvature directions to vary with $\theta$ and differ for train and test. Since (\ref{eq:taylor_test_loss}) is a local approximation, the argument above still holds, and we find that the local zero-curvature directions do not affect the test performance near a train local minimum. However, (\ref{eq:sgdss}) is more complicated, because it depends on the global landscape where the set of nonzero-curvature directions can vary (also, the gradient variance only has rank $N-1$, causing technical difficulties), which is why we require the problem modifications discussed in Section \ref{sec:overparam}. In addition to $\ell_2$-regularization on the train loss, we also need to add a small amount ($\beta)$ of noise to the diffusion matrix to make it invertible, which changes the problem so that SGD does not force the zero-curvature directions to zero as in the discussion above, but instead converges to an infinite-support distribution, with the (local) zero-curvature directions represented locally by Gaussians with mean zero and variance $T\beta/\alpha$.

\subsection{Nonzero-Curvature Subspaces}
\label{append:overparam_remarks}

Our goal is to derive the results in Section \ref{sec:overparam}, but we first take a detour to describe in detail why and how overparametrization changes the problem.
In the underparametrized case (more data samples $N$ than model parameters $p$), the per-sample gradients $\partial_\theta U(x_i, \theta)$ typically span all of $\RR^p$, while in overparametrized case ($p > N$), the per-sample gradients span a subspace of dimension no larger than $N$. To work out the many consequences of this, we introduce some more compact notation. We consider a typical loss function
$U(\theta) = \frac{1}{N} \sum_{i=1}^N U(x_i, \theta).$
The test loss and the train loss without $\ell_2$-regularization in (\ref{eq:losses}) both have this form. Let
$ g_i(\theta) = \partial_\theta U(x_i, \theta) \in \RR^{p}$
be a shorthand for the $N$ per-sample gradients, and
$\mathcal{G}(\theta) = \text{span}\{ g_i(\theta) \}_{i=1, \ldots, N}.$ Thus $\mathcal{G}(\theta)$ is the nonzero-curvature subspace at $\theta$.
Let $\mathcal{G}^\perp = \RR^p \backslash \mathcal{G}_\textit{tr}$ denote the complement subspace orthogonal to $\mathcal{G}$. To simplify exposition, we assume that all per-sample gradients are linearly independent, so that the dimension of $\mathcal{G}$ is $\min(p, N)$. Thus, if $N \ge p$, then $\mathcal{G} = \RR^p$ and $\mathcal{G}^\perp$ is empty; but if $p > N$, then $\mathcal{G}$ has dimension $N$ and $\mathcal{G}^\perp$ has dimension $p - N$. Because the gradients are generally functions of $\theta,$ the spaces $\mathcal{G}$ and $\mathcal{G}^\perp$ can change with $\theta.$ 

As shown at the end of this section, the average gradient is in $\mathcal{G}$, the range of the gradient covariance is contained in $\mathcal{G}$ but has one less dimension, and the range of the Hessian is equal to $\mathcal{G}$ provided $\mathcal{G}$ is constant in a neighborhood around $\theta$. In particular, the gradient covariance and Hessian are both rank-deficient (hence non-invertible). All of these facts apply directly to the test loss, and to the part of the train loss excluding $\ell_2$-regularization. Adding $\ell_2$-regularization to the train loss changes the results as follows. The average gradient is $\partial_\theta U^\textit{tr}(\theta) = (1/N) \sum g_i(\theta) + \alpha \theta$, so that $\partial_\theta U^\textit{tr}(\theta)$ is no longer constrained to lie in $\mathcal{G}^\textit{tr}$ and can be anywhere in $\RR^p$. The gradient covariance is unchanged since $\alpha \theta$ is not random; in particular, it is still rank-deficient. The Hessian $\partial_\theta^2 U^\textit{tr}(\theta)$ becomes full-rank, because regularization adds $\alpha$ to every eigenvalue of the Hessian. These facts have important implications for the Taylor approximation (\ref{eq:taylor_test_loss}) and the SGD steady-state (\ref{eq:sgdss}) in the overparametrized case. First, recalling that a local minimum is strict when the Hessian is full-rank, we see that test local minima are non-strict, while train local minima are strict only if we add $\ell_2$-regularization. Second, $D^\textit{tr}(\theta)$ is non-invertible, invalidating (\ref{eq:sgdss}), but we can resolve this by adding isotropic Gaussian noise to the SGD updates to make $D^\textit{tr}(\theta)$ full rank, together with $\ell_2$-regularization to control the part of the drift term in $\mathcal{G}_\textit{tr}^\perp$.

\subsubsection*{Average gradient, covariance, and Hessian are in $\mathcal{G}$}
The average gradient is
$ g(\theta) = \frac{1}{N} \sum_{i=1}^{N} g_i(\theta),
$
which clearly lies in the span of the per-sample gradients: $g(\theta) \in \mathcal{G}(\theta).$ 
The gradient covariance is
$\text{Cov}(g(\theta)) = \frac{1}{N} \sum_{i=1}^{N} g_i g_i^T - g g^T \in \RR^{p \times p}.$
We will show shortly that the range of $\text{Cov}(g)$ is also contained in $\mathcal{G}$, but it is not equal to it, since $\mathcal{G}$ has one more dimension. This means that $\text{Cov}(g)$ has a nonempty nullspace containing $\mathcal{G}^\perp$ and an extra direction, hence $\text{Cov}(g)$ is not invertible. To see that $\text{range}(\text{Cov}(g)) \subset \mathcal{G}$, note that $\text{Cov}(g)$ (like any $p \times p$ sample covariance constructed from $N$ data points) has rank $\text{min}(p, N-1)$ (since it can be written as a sum of $N$ rank-1 matrices, $\sum (g_i-g)(g_i-g)^T$, but $\text{span}\{ g_i - g\}$ only has dimension $N-1$ when $p > N$). Therefore, when $p > N$, the nullspace of $\text{Cov}(g)$ has dimension $p - N + 1$. It is also clear that the nullspace of $\text{Cov}(g)$ contains $\mathcal{G}^\perp$ (to see this, consider any vector $w \in \mathcal{G}^\perp$, i.e., such that $w^T g_i = 0$ for all $i \in \{1, \ldots , N\}$; it then follows from the definition of $\text{Cov}(g)$ that $\text{Cov}(g) w = 0$). Therefore, the nullspace of $\text{Cov}(g)$ contains all of $\mathcal{G}^\perp$ (which has dimension $p-N$), plus one extra direction. Equivalently, the range of $\text{Cov}(g)$ is contained in $\mathcal{G}$, but $\mathcal{G}$ has one extra direction. 
Finally, the range of the Hessian $\partial_\theta^2 U(\theta)$ is equal to $\mathcal{G}(\theta)$ if $\mathcal{G}(\theta)$ is constant in some open neighborhood of $\theta$. To see this, note that for any $w \in \mathcal{G}^\perp(\hat \theta)$, we have that
$\partial_\theta^2 U(\hat \theta) w = \partial_\theta (g(\theta)^T)|_{\hat \theta} w = \partial_\theta (g(\theta)^T w)|_{\hat \theta} = 0$, since $w \in \mathcal{G}^\perp(\theta) \implies g(\theta)^T w = 0$ for all $\theta$ in a neighborhood of $\hat \theta$ by assumption. So the nullspace of $\partial_\theta^2 U(\hat \theta)$ contains $\mathcal{G}^\perp(\hat \theta)$. Also, assuming that the per-sample Hessians in the dataset are linearly independent, the range of the Hessian has dimension $N,$ so it is equal to $\mathcal{G}$. 

\subsection{Overparametrized SGD Steady-State}
\label{append:ss_overparam}
In the overparametrized case, we use the modified SGD (\ref{eq:SGDMod}) which yields the diffusion approximation with the same FP description (\ref{eq:fp}) but with modified probability current $J(\theta)= \big(\partial_{\theta} U^\textit{tr}(\theta) + \alpha \theta \big)\rho(\theta) + \frac{T}{2} \partial_\theta \cdot \big( (D^\textit{tr}(\theta) + \beta^2 I) \rho(\theta) \big).$ 
Recall that the diffusion matrix in the diffusion process approximation is the covariance of $\Delta_t$, which is now $\lambda T (D^\textit{tr}(\theta) + \beta^2 I),$ and is invertible when $\beta \neq 0$. But the added isotropic noise creates a new problem: since $\partial_\theta U(x_i, \theta)$ lies in $\mathcal{G}_\textit{tr},$ we now have diffusion noise but no drift in $\mathcal{G}_\textit{tr}^\perp$, creating the possibility of a distribution of parameters that spreads out forever as $t \to \infty$. To prevent this, we need drift in $\mathcal{G}_\textit{tr}^\perp$ that keeps parameter values small, and $\alpha > 0$ accomplishes this for us. 
So, with $\alpha, \beta > 0$, the same approach given in Section \ref{append:steady-state} works to obtain the steady-state distribution, resulting in the same form after the appropriate redefinitions of the diffusion matrix (adding $\beta^2I$ to it) and drift term (adding $\alpha \theta$ to it). With $\beta=0$, it is unclear how find the steady-state distribution: for example, one might try replacing the inverse of $D^\textit{tr}$-projected-onto-$\mathcal{G}_\textit{tr}$ with its pseudoinverse, and including a delta function in $\rho$ to enforce conditions implied by the ODE in the one-dimensional nullspace of $D^\textit{tr}$-projected-onto-$\mathcal{G}_\textit{tr}$, but this choice turns out to yield a nonzero divergence of the probability current $\nabla \cdot J \neq 0$, hence it is not a steady-state solution.

\subsection{Overparametrized Taylor Approximation of Test Loss}
\label{append:taylor_overparam}

If the local minima of train and test are strict (i.e. single points), then the Taylor expansion approximation of the test loss (\ref{eq:taylor_test_loss}) goes through unchanged. As we saw in Section \ref{append:overparam_remarks}, the local minima of train are always strict since we assume that the train loss includes nonzero $\ell_2$-regularization (which makes the Hessian full-rank). However, the local minima of test are non-strict in the overparametrized case since the Hessian is rank-deficient (unless we also add $\ell_2$-regularization to the test loss, which is not usually done in practice). However, we can still find a natural analog of the Taylor approximation in the overparametrized case. The test local minima are no longer strict. We assume that $\mathcal{G}_\textit{test}$ is constant in a neighborhood of $\bar{\theta}_k^\textit{test}$, where $g^\textit{test}(\bar{\theta}_k^\textit{test}) = 0$. From Section \ref{append:overparametrization}, we know that $\partial_\theta U^\textit{test}(\theta) = g^\textit{test}(\theta)$, where $g^\textit{test}(\theta) \in \mathcal{G}^\textit{test}(\theta)$, and $\mathcal{G}^\textit{test}(\theta)$ has dimension $N < p$. Therefore (as shown at the end of this section) each test local minimum is a set:
\begin{align}
\theta^{\textit{test}}_k = \{ \theta^{\textit{test}}_k(w) \equiv \text{Proj}_{\mathcal{G}_\textit{test}} (\bar{\theta}_k^\textit{test}) + w , \text{ for any } w \in \mathcal{G}_\textit{test}^\perp \}, \text{ where } g^\textit{test}(\bar{\theta}_k^\textit{test}) = 0. \label{eq:test_loc_min_set}
\end{align}
The local minima of the train loss are strict due to the $\ell_2$-regularization. (The train gradients are $\partial_\theta U^\text{train}(\theta) = g^\text{train}(\theta) + \alpha \theta$, which can only be zero when the element in $\mathcal{G}_\textit{tr}^\perp$ is equal to zero). Recall that in (\ref{eq:taylor_test_loss}), we seek to approximate the test loss at a point $\hat \theta$ found by a single trajectory of SGD. We know from the previous section that $\hat \theta$ must have its projection on $\mathcal{G}^\perp$ equal to zero, so it is also unique. Taylor-expanding the test loss about any $\theta^{\textit{test}}_k(w)$ in the test-loss-local-minimum subspace, we obtain:
\begin{align*}
    U^\textit{test}(\theta^\textit{tr}_k) &\approx U^\textit{test}(\theta^{\textit{test}}_k(w)) + \half (s(w) + \hat b)^T \partial_\theta^2 U(\theta^{\textit{test}}_k(w)) (s(w) + \hat b), \quad \text{for any $w$} \\
    &= U^\textit{test}_k + \half (s(w) + \hat b^T C^\textit{test}_k (s(w) + \hat b), \quad \text{since $U$, $\partial_\theta^2 U$ do not depend on $w\in \mathcal{G}_\textit{test}^\perp$}, \\
    \text{where } s(w) &= \theta^{\textit{test}}_k(w) - \theta^\textit{tr}_k, \quad \text{and} \quad
    \hat b = \hat \theta - \theta^\textit{tr}_k.
\end{align*}
Since the range of $C^\textit{test}_k$ is orthogonal to $\mathcal{G}_\textit{test}^\perp$, the term $s(w)^T C^\textit{test}_k s(w)$ only depends on the part of $s(w)$ in $\mathcal{G}_\textit{test}$, hence the Taylor approximation of $U^\textit{test}(\theta^\textit{tr}_k)$ is the same for any choice of $w$. We would get the same result for $U^\textit{test}(\theta^\textit{tr}_k)$ by defining $s_k = s(w)$ for any $w$, but a natural choice (that eliminates the irrelevant-to-the-test-loss part of $s$ in $\mathcal{G}_\textit{test}^\perp$) is
$ s_k = \text{Proj}_{\mathcal{G}_\textit{test}} \big( \theta^{\textit{test}}_k - \theta^\textit{tr}_k \big). $
(This choice of $s(w)$ corresponds to $w = \text{Proj}_{\mathcal{G}_\textit{test}^\perp}(\theta^\textit{tr}_k)$, which is the the solution to $\min_w \| s(w) \|$.)
The overall result is:
\begin{align*} 
    U^\textit{test}(\hat \theta) &= U^\textit{test}_k + \half (s_k + \hat b)^T C_k (s_k + \hat b), 
    \text{ with } s_k = \text{Proj}_{\mathcal{G}_\textit{test}} \big( \theta^{\textit{test}}_k - \theta^\textit{tr}_k \big)
    \text{ and } \hat b = \hat \theta - \theta^\textit{tr}_k.
\end{align*}
We see that the Taylor approximation of the test loss at a train local minimum has the same form as in Equation \ref{eq:taylor_test_loss}, but the definition of the shift takes a more general form: it is now the projection of the difference between the test and train local minima onto $\mathcal{G}_\textit{test}$ (subspace for the test loss). This definition coincides with the underparametrized definition of $s_k$ when $\mathcal{G}_\textit{test}$ is all of $\RR^p$. 

\subsubsection*{Test local minima are given by (\ref{eq:test_loc_min_set})}
Let $\mathcal{O} = \begin{bmatrix} \mathcal{O}_{\mathcal{G}} & \mathcal{O}_{\mathcal{G}^\perp} \end{bmatrix}$ be an orthonormal matrix such that $\mathcal{O}_{\mathcal{G}}$ is a basis for $\mathcal{G}_{\textit{test}}$ and $\mathcal{O}_{\mathcal{G}^\perp}$ is a basis for $\mathcal{G}_{\textit{test}}^\perp$. So, i.e., the projector onto $\mathcal{G}_{\textit{test}}$ is $\text{Proj}_{\mathcal{G}_\textit{test}}=\mathcal{O}_{\mathcal{G}}\mathcal{O}_{\mathcal{G}}^T.$ Define 
$z = \mathcal{O}^T \theta = \begin{bmatrix} \mathcal{O}_{\mathcal{G}}^T \theta \\ \mathcal{O}_{\mathcal{G}^\perp}^T \theta \end{bmatrix} \equiv \begin{bmatrix} z_{\mathcal{G}} \\ z_{\mathcal{G}^\perp} \end{bmatrix}.$
Then
\begin{align*}
    \begin{bmatrix} \partial_{z_\mathcal{G}} U^\textit{test}(\theta) \\ \partial_{z_{\mathcal{G}^\perp}} U^\textit{test}(\theta) \end{bmatrix} &\equiv \partial_z U^\textit{test}(\theta) 
    = \partial_z U^\textit{test}(\mathcal{O}z)
    = \mathcal{O}^T \partial_\theta U^\textit{test}(\theta) = \begin{bmatrix} \mathcal{O}_\mathcal{G}^T g(\theta) \\ 0 \end{bmatrix}.
\end{align*}
In particular, we have
$ \partial_{z_{\mathcal{G}^\perp}} U^\textit{test}(\theta) = 0.$ Therefore the unregularized train loss must be constant w.r.t. $z_{\mathcal{G}^\perp}$, so we can write (by picking $z_{\mathcal{G}^\perp} = 0$)

\begin{align}
    U^\textit{test}(\theta) = U^\textit{test}(\mathcal{O} z) &= U^\textit{test} \biggr( \begin{bmatrix} \mathcal{O}_{\mathcal{G}} & \mathcal{O}_{\mathcal{G}^\perp} \end{bmatrix} \begin{bmatrix} z_{\mathcal{G}} \\ 0 \end{bmatrix} \biggr)
    = U^\textit{test}(\mathcal{O}_{\mathcal{G}} z_{\mathcal{G}})
    = U^\textit{test}(\text{Proj}_{\mathcal{G}_\textit{test}}(\theta)) \notag\\
    \implies U^\textit{test}(\theta^{\textit{test}}_k(w)) &= U^\textit{test}(\text{Proj}_{\mathcal{G}_\textit{test}} (\bar{\theta}_k^\textit{test})) = U^\textit{test}(\bar{\theta}_k^\textit{test}) \notag\\
    \implies \partial_\theta U^\textit{test}(\theta^{\textit{test}}_k(w)) &= \partial_\theta U^\textit{test}(\bar{\theta}^{\textit{test}}_k(w)) = 0.
\end{align}


\section{Higher Order Curvature}
\label{append:higher_order}

The full Taylor-expansion of a function $U(\theta)$ (for example, the train or test loss) about a local minimum $\theta_k$ is given by:
\begin{align*}
    U(\theta) &= U(\theta_k) + \sum_{j=2}^\infty \frac{1}{j!} \partial_\theta^j U(\theta_k) (\theta - \theta_k)^j 
\end{align*}
Here we are using multi-index notation, so that
\begin{align*}
    \frac{1}{j!} \partial_\theta^j U(\bar \theta) \theta^j &\equiv
    \sum_{j: j_0 + \ldots + j_{n-1} = |j|} 
    \frac{1}{j_0! j_1! \ldots j_{n-1}!} 
    \frac{\partial^{|j|} U(\bar \theta) }{\partial^{j_0}_{\bar \theta} \partial^{j_1}_{\theta_1} \ldots \partial^{j_{n-1}}_{\theta_{n-1}}} \bar \theta^{j_0} \theta_1^{j_1} \ldots \theta_{n-1}^{j_{n-1}}.
\end{align*}
Note that the derivatives of order 3 and higher are tensors, and $\partial_\theta^j U(\theta_k) \theta^j$ represents a tensor contraction.

In one dimension, a critical point $\theta_k$ of a function $f: \RR \to \RR$ is a local minimum if 
\begin{align}
J = \argmin \{ j: \partial_\theta^{j} f(\theta_k) \neq 0 \} \text{ is even, and } \partial_\theta^{J} f(\theta_k) > 0.
\label{eq:loc_min_1d}
\end{align}
This follows from Taylor's theorem. For example, if $\partial_\theta^2 f(\theta_k) = 0$, then we can look at the next two derivatives:if  $\partial_\theta^3 U(\theta_k) \neq 0$ then $\theta_k$ is a saddle point; otherwise, it is a local minimum if $\partial_\theta^4 f(\theta_k) > 0$ and a local maximum if $\partial_\theta^4 f(\theta_k) < 0$. If the 3rd and 4th derivatives are zero, we need to look at the 5th and 6th, and so on. In higher dimensions ($U: \RR^p \to \RR$), we can generalize $\partial_\theta^{J} f(\theta_k) > 0$ to $\partial_\theta^{J} f(\theta_k) x^J > 0$ for all $x$ (analogous to definition of positive definite Hessian). But the condition for a local minimum becomes more complex since, for example, the second derivative could be positive semidefinite (with both positive and zero eigenvalues); in positive-eigenvalue directions we can be confident of a local minimum, but in the nullspace we must consider higher derivatives. However, we also know that a critical point $\theta_k$ is a local minimum of $U$ if $\theta_k$ is a local minimum of the 1D function obtained by evaluating $U$ along \emph{any line} through $\theta_k$. This characterization allows us to fall back to the simpler 1D characterization (\ref{eq:loc_min_1d}) along any line, and suffices for many of our purposes. 

In particular, in the Taylor expansion (\ref{eq:taylor_test_loss}), taking $\hat \theta = \theta^\textit{tr}_k$ to simplify the argument, we wish to approximate $U^{\textit{test}}(\theta^\textit{tr}_k)$ by expanding about $\theta^\textit{test}_k$, so we only care about the values of $U^{\textit{test}}$ along the line between the train and test local minima. Defining:
\begin{align}
    f^{\textit{test}}(r) &= U^{\textit{test}}(\Theta(r)), \quad \Theta(r) 
    = \theta_k^{\textit{tr}} + \frac{r s_k}{\|s_k\|}
    \implies \Theta(0) = \theta_k^{\textit{tr}}, \text{ } \Theta(\|s_k\|) = \theta_k^{\textit{test}} \nonumber \\
    \implies U^{\textit{test}}(\theta^\textit{tr}_k) &= f^{\textit{test}}(0) = f^{\textit{test}}(\|s_k\|) + \sum_{j=2}^\infty \frac{(-\|s_k\|)^{|j|}}{j!} \partial_r^j f^{\textit{test}}(\|s_k\|), \nonumber\\
    \text{with } &\partial_r^j f^{\textit{test}}(r) = \partial_\theta^j U^{\textit{test}}(\Theta(r)) \biggr(\frac{s_k}{\|s_k\|}\biggr)^j \label{eq:loss_along_a_line}
\end{align}
(Note that this Taylor expansion looks slightly unusual because it is \emph{centered} at $\|s_k\|$, the test local minimum, and \emph{evaluated} at 0, the train local minimum, so that $dr = -\|s_k\|$.) Since $\theta^{\textit{test}}_k$ is a local minimum of $U^{\textit{test}}$, we know that $f^\textit{test}$ has a local minimum at $\|s_k\|$, so condition (\ref{eq:loc_min_1d}) must hold there. This means that the order $J$ of the minimum nonzero derivative along the line must be even, and we must have:
$$\partial_r^J f^{\textit{test}}(\|s_k\|) = \partial_\theta^J U^\textit{test}(\theta_k^{\textit{test}}) \biggr(\frac{s_k}{\|s_k\|}\biggr)^J > 0. $$
This offers a way to characterize the curvature in terms of a higher-even-order derivative $\partial_\theta^J U^\textit{test}$, even if the Hessian $\partial_\theta^2 U^\textit{test}$ is zero along the line between the local minima. Higher curvature along this line corresponds to larger values of $\partial_\theta^J U^\textit{test}(\theta_k^{\textit{test}}) s_k^J$.
We can also partially generalize the SGD steady-state results to the higher-order curvature case. However, we currently only know how to evaluate the necessary integrals in the special case where the local curvature of $v$ depends \emph{only on a single even derivative} (for example, only a positive-definite 4th derivative), and where the even derivative tensor is \emph{diagonalizable} (which is not the case in general for tensors of order 3 and higher).
That is, we must assume that at its local minima $v$ can be approximated as:
\begin{align*}
   v(\theta) &\approx v(\mu_k) + \frac{1}{J!} \partial_\theta^J v(\mu_k) (\theta - \mu_k)^J,
\end{align*}
where $J$ is even, and $\partial_\theta^J v(\mu_k)$ is diagonalizable. Then
\begin{align*}
    \rho(\theta) &\propto \text{exp}\{-\frac{2}{T} v(\theta)\} \approx \text{exp}\{-\frac{2}{T} v(\mu_k)\} \text{exp}\{-\frac{2}{T J!} \partial_\theta^Jv(\mu_k) (\theta - \mu_k)^J \}
\end{align*}
To find the basin weights, we need to evaluate the integral:
\begin{align*}
    \int_{\mathcal{B}_k} \rho(\theta) d\theta &\propto \text{exp}\{-\frac{2}{T}v(\mu_k)\} \int_{\mathcal{B}_k} \text{exp}\{-\frac{2}{TJ_k!} \partial_\theta^{J_k} v(\mu_k) (\theta - \mu_k)^{J_k} \} d\theta \\
    &= \text{exp}\{-\frac{2}{T}v(\mu_k)\} \biggr(\frac{TJ_k!}{2}\biggr)^{\frac{n}{|J_k|}} \int e^{-\partial_\theta^j v(\mu_k) y^j} dy \label{eq:basin_integral}
\end{align*}
To make more progress, we need to evaluate integrals of the form 
$\int_{-\infty}^\infty \text{exp}\{-\partial_x^j v(\mu) x^j \} d\theta$ (here $j$ is even but otherwise arbitrary).
This is where we need the assumption that $\partial_x^j v(\mu)$ is diagonalizable , which always holds for symmetric tensors of order two with real entries (i.e.\ matrices), but does not hold in general for symmetric tensors of higher orders (\cite{comon1994tensor}). Specifically, we assume that $\partial_x^j v(\mu)$ can be written as:
\begin{align*}
    \partial_x^j v(\mu) &= \sum_{i=1}^r \lambda_i q_i^{\otimes j}, \quad \text{with $q_i$'s orthogonal, } \quad \lambda_i > 0, \quad r \le n \\
    \implies 
    \partial_x^j v(\mu) x^j &= \sum_{i=1}^r \lambda_i q_i^{\otimes j} x^j 
    = \sum_{i=1}^r \lambda_i (q_i^T x)^j 
    = \sum_{i=1}^r \lambda_i z_i^{j}, \quad \text{with } z = Q^Tx, \quad Q = \begin{bmatrix} v_1 \ldots v_n \end{bmatrix},
\end{align*}
where if $r < n$ we add $n-r$ additional columns to $Q$ to complete an orthonormal basis for $\RR^n$. (Note that the assumption that the $q_i$'s are orthogonal implies that $r \le n$ since more than $n$ vectors in $\RR^n$ can't be linearly independent, and also implies that $\lambda_i \ge 0$ for all $i$ because $\partial_x^j v(\mu)$ is positive semidefinite (\cite{qi2005eigenvalues}). If we take only the first $r$ nonzero eigenvalues we can assume $\lambda_i > 0$ for all $i=1, \ldots, r$. In this case we have an analog of the determinant given by \cite{qi2005eigenvalues}: $ |\partial_x^j v(\mu)| = \prod_{i=1}^r \lambda_i. $
Then:
\begin{align*}
    \int_{-\infty}^{\infty} e^{-\partial_x^j v(\mu) x^j} dx &= \int_{-\infty}^{\infty} e^{-\sum_{i=1}^r \lambda_i z_i^{j}} dz, \quad z = Q^Tx, \quad dx = |Q| dz = dz\\
    &= \prod_{i=1}^r \int_{-\infty}^{\infty} e^{-\lambda_i z_i^{j}} dz_i 
    = \prod_{i=1}^r \lambda_i^{-\frac{1}{j}} \int_{-\infty}^{\infty} e^{-y_i^{j}} dy_i, \quad y_i = \lambda_i^{1/j} z_i \\
    &= |\partial_x^j v(\mu)|^{-\frac{1}{j}} \int_{-\infty}^{\infty} e^{-s^{j}} ds, 
    = 2 |\partial_x^j v(\mu)|^{-\frac{1}{j}} \Gamma\biggr(\frac{j+1}{j}\biggr) \\
    \implies \int_{\mathcal{B}_k} \rho(\theta) d\theta &\approx \frac{\text{exp}\{-\frac{2}{T}v(\mu_k)\}}{|\partial_x^{J_k} v(\mu)|^{\frac{1}{J_k}}} \biggr(\frac{TJ_k!}{2}\biggr)^{\frac{n}{J_k}} 2  \Gamma\biggr(\frac{J_k+1}{J_k}\biggr)
\end{align*}
To see that agrees with the $J = 2$ case, note that
$\Gamma(3/2) = \frac{\sqrt{\pi}}{2}.$

\section{Experiment Notes}
\label{append:expt_notes}
\subsection*{Shift-curvature symmetrized quadratic fits}
Due to the shapes of the losses we find that the symmetrized quadratic fit more accurately captures the relevant overall curvature of the basin. First, the loss curves tend to have a small `flattened' region right around the local minima, despite looking roughly quadratic overall (we suspect the `flattening' is due to sigmoid saturation), so that the curvature evaluated right at the local minima tends to underestimate the overall curvature of the basin, which is better captured by fitting a quadratic. Second, we find that the train and test losses have an asymmetrical shape (\cite{he2019asymmetric}) along the line between the local minima, hence we reflect about the test local minimum in the direction toward the train local minimum since this is the direction that affects the test loss at the test local minimum; similarly, we reflect the train loss about its local minimum in the same direction (i.e. away from the test local minimum) since we are interested in train curvature as a proxy for test curvature so we should study the same direction of each. We believe the observed asymmetry of the losses is due to the experimental design: to find the test minimum, we run gradient descent starting from close to the train minimum, which means we go down the steepest direction of the test basin away from the train local minimum until we reach a test local minimum; however if the test basin has a flat region at the bottom, continuing in the same direction may have zero or very small test loss until we reach the opposite wall of the basin (however, this direction has little impact on the test loss at the train local minimum).

\section{Additional Experiments}

\subsection{Temperature and Shift Experiments}
\label{append:temp_expt}

\begin{figure}[t]
\centering
\includegraphics[width=0.49\columnwidth]{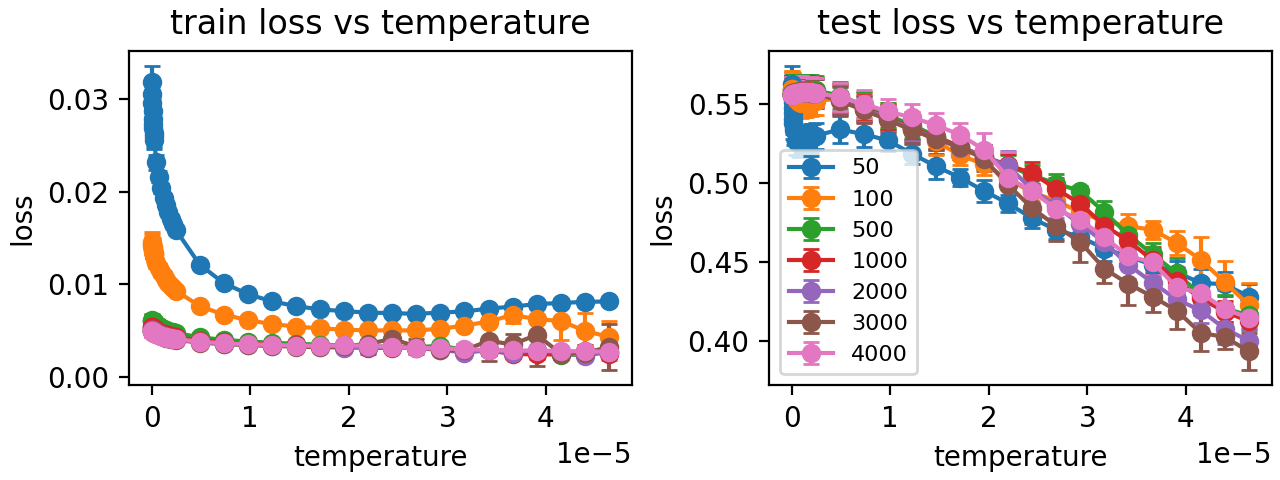}
\includegraphics[width=0.24\columnwidth]{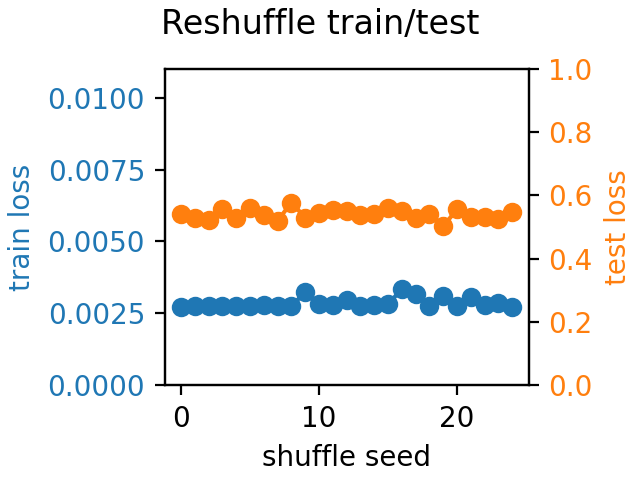}
\includegraphics[width=0.24\columnwidth]{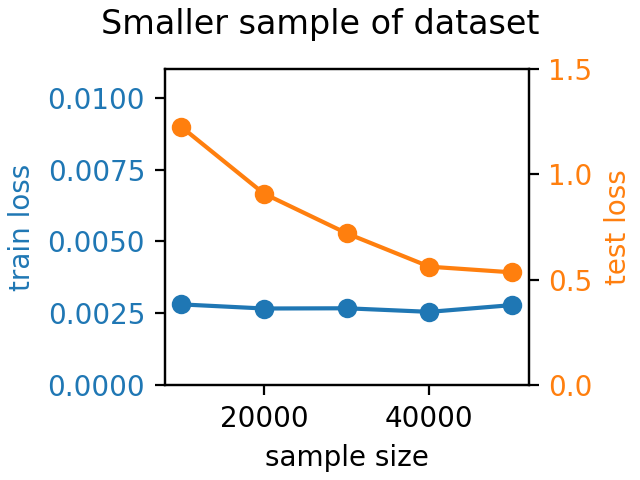}\\
\caption{(Left two plots) Temperature experiment. Plots show the train and test loss on CIFAR10 as a function of temperature (learning rate divided by batch size). Different colors correspond to different batch sizes. The train loss is consistently close to zero (except for smallest batch sizes, as discussed in the text), while the test loss improves with increasing temperature in a consistent way for all batch sizes; suggesting that temperature is the most important variable (rather than batch size or learning rate).
(Right two plots) Shift experiments (shuffling and sample-size). Reshuffling the training and test sets has little impact on the final train/test losses, suggesting that there is no distribution shift between the default train and test sets. In contrast, smaller subsamples of the dataset (still proportionally split into train/test) lead to higher test losses while train loss stays small, suggesting that finite sampling is the source of train/test shift.
}
\label{fig:vgg_lb_init_hpuh5hk7ja}
\end{figure}


Figure \ref{fig:vgg_lb_init_hpuh5hk7ja} shows experiments on CIFAR10 (\cite{krizhevsky2009learning}) to demonstrate the phenomenon of generalization improving with temperature, as well as to understand the shift between and training and test sets for CIFAR10. We used the VGG9 network \cite{simonyan2014very} and the training procedure of \cite{li2017visualizing} (but with no momentum).
For the temperature experiment, we first trained a network using a large batch size (4096) and decreasing learning rate schedule (starting from 0.1, scaled by 0.1 at 0.5, 0.75 of total epochs, ending with LR 0.001) until the training converged. Then, we continued training from that initialization with a variety of batch sizes and learning rates (all LRs held constant in second stage), for 1000 epochs (regardless of batch size). For each second-stage run, we took the median of the last 100 steps (to eliminate spiky outliers) as the `final loss'. We repeated this two-stage experiment 10 times (with different initializations), and plotted the mean and standard deviation over all trials of the final train and test loss as a function of temperature (learning rate divided by batch size) (from 2.44e-8 to 2.44e-5). (At end of first stage, final temperature = 2.44e-7; mean final train loss = 0.005 (var 4e-07) and test loss = 0.555 (var 1e-4).)
Our results show that both the train loss and test loss depend primarily on the temperature, and test loss decreases with increasing temperature while train loss remains roughly constant. For small batches (50 and 100), train loss was higher at the beginning of the second stage for all LRs; train loss dropped significantly by the end of training with large LRs but remained high with small LRs. We believe this was due to mismatched batchnorm statistics and parameters (parameters learned for batch size 4096 but applied to batch size 50 or 100); with larger LRs the network was able to retrain the batchnorm parameters, but smaller LRs did not allow sufficient progress. 

The shift experiments aimed to clarify whether the train/test shift in CIFAR10 is more likely due to distribution shift or to finite sampling. The training procedure was the same as in the initial stage of the temperature experiment. To study distribution shift, we ran a shuffling experiment, where we merged the train and test sets, reshuffled them to create a single dataset, and then randomly split the data into training and test at each trial (25 total). If there were a distribution shift between the default train and test split in the CIFAR data, this process would remove it, and we would expect the test loss to decrease. However, our experiment shows that there is no significant difference in the test loss for different shufflings of the train and test sets, suggesting that distribution shift is not present.
To better understand the role of finite sampling, we ran a sample-size experiment, where for each trial we chose a subsample of the full dataset of a given size, split it proportionally into train and test, and trained on the training set. If finite sampling were causing the shift between the training and test distributions, we would expect smaller sample sizes to exacerbate the difference (since sampled distributions generally become closer to the underlying distribution as the sample size grows). Our experiment shows that the test loss increases as the sample size decreases, consistent with finite sampling as the source of distribution shift.

\subsection{One and Three Basin Experiments}
\label{append:more_synth_exp}

\begin{figure}[t]
\centering
\includegraphics[width=0.45\columnwidth]{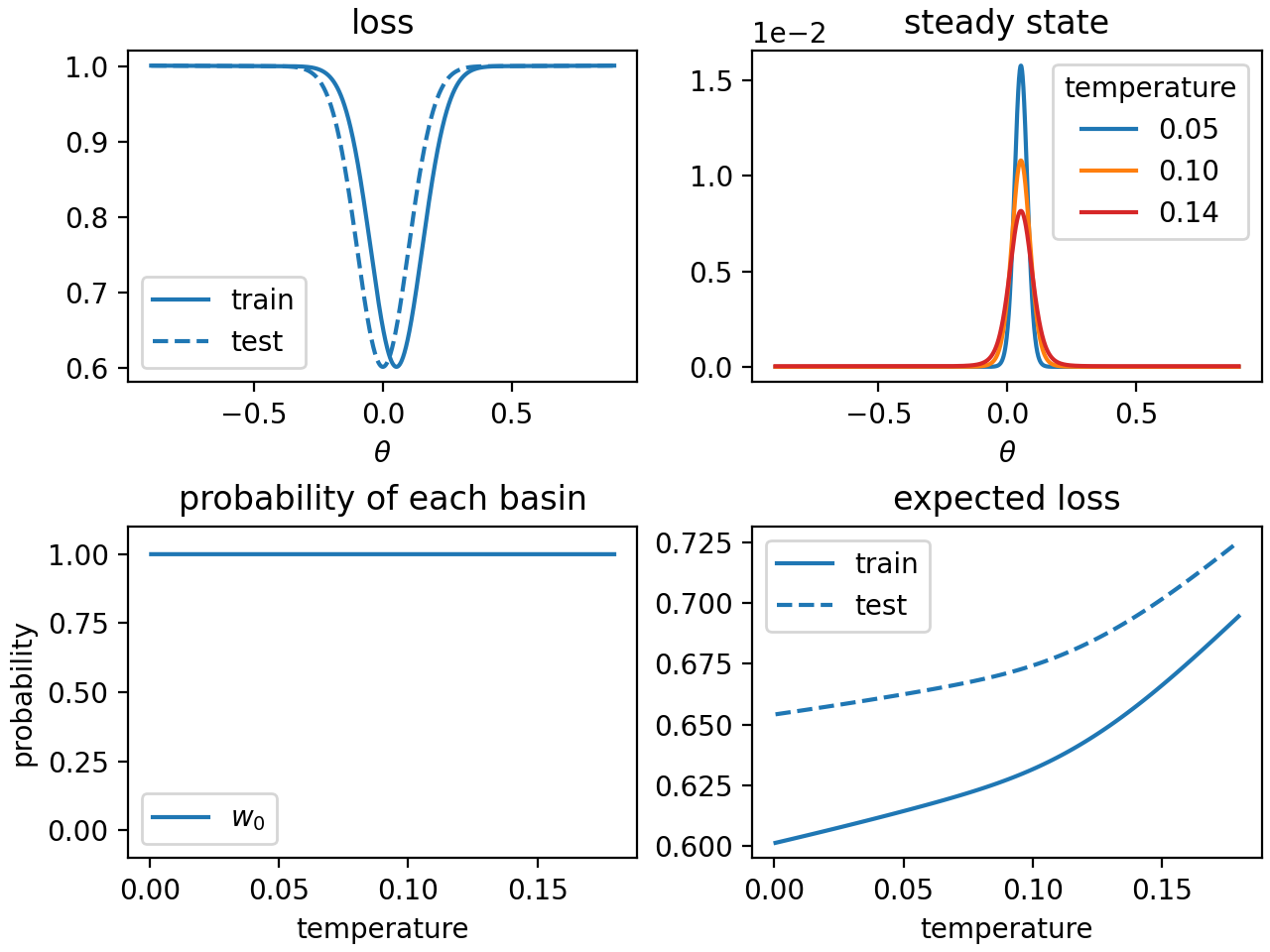} \quad \quad
\includegraphics[width=0.45\columnwidth]{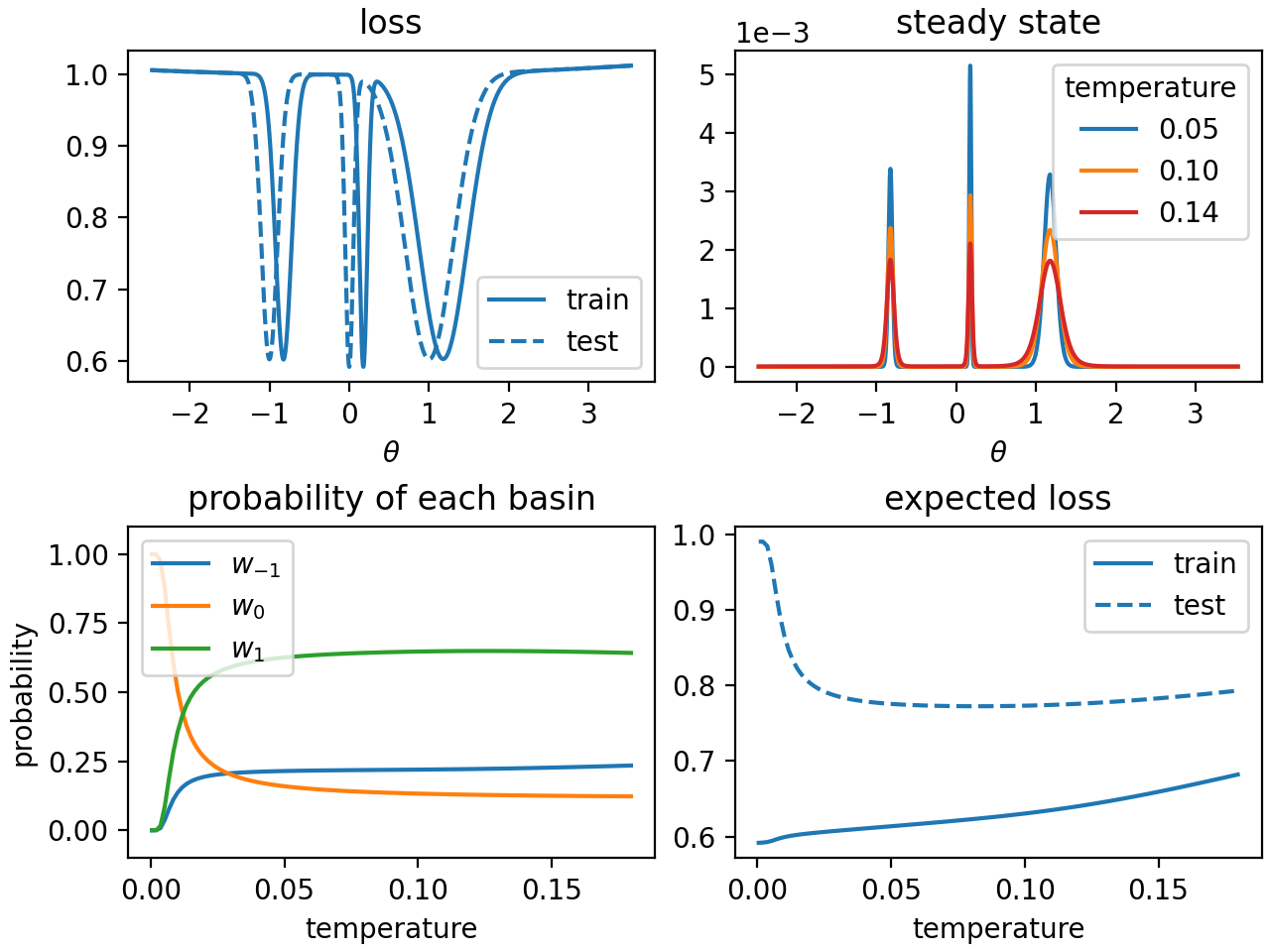}\\
\caption{
Left: Synthetic one-basin experiment with the same setup as in Figure \ref{fig:2basin_g_A}. Train and test error both increase as temperature increases.
Right: Synthetic three-basin experiment with the same setup as in Figure \ref{fig:2basin_g_A}. Minimum at $0$ is deeper but narrower than minima at $-1$ and $1$; $-1$ and $1$ are equally deep but $1$ is wider. Generalization improves (test error decreases while training error increases) with increasing temperature, as the steady-state distribution moves probability mass from the narrower deeper minimum at $0$ toward the wider minima, with a preference for the widest minimum at $1$.
}
\label{fig:1_3_basin}
\end{figure}

Figure \ref{fig:1_3_basin} shows 1-basin and 3-basin experiments analogous to the 2-basin experiment in Figure \ref{fig:2basin_g_A}. In the 1-basin example, both train and test error increase with increasing temperature. The 3-basin example shows qualitatively similar behavior to the 2-basin case and confirms the SGD steady-state distribution's preference for wider minima when the depths are similar.

\subsection*{Experiment Details and Hyperparameters}
\subsubsection*{Figures \ref{fig:train_test_loss}, \ref{fig:vgg_shift_curv}, \ref{fig:resnet_nbn_shift_curv}}
\begin{itemize}
    \item Resnet10 model: \cite{shon2021resnet}'s \verb|haiku| implementation of Resnet16 (\cite{he2016deep}). Modified by reducing blocks per group from 3 to 2, and removing batch norm.
    \item VGG16 model: \cite{yoshida2021vgg}'s \verb|haiku| implementation of VGG16 (\cite{simonyan2014very}). Adapted for CIFAR10 by reducing size of dense layers to 512 as proposed by \cite{liu2015very} and removing adaptive pooling.
    \item Data: CIFAR10 (\cite{krizhevsky2009learning}).
    \item Training: adapted from \cite{shon2021resnet}'s \verb|JAX| sample code.
\end{itemize}

\begin{small}
\begin{tabular}[t]{l}
    Initial SGD \\
    \verb|BATCH_SIZE: 100|\\
    \verb|TRAIN_SAMPLE_COUNT: 50000|\\
    \verb|TRAIN_SHUFFLE: True|\\
    \verb|TRAIN_DROP_LAST: True|\\
    \verb|MAX_EPOCH: 400|\\
    \verb|MAX_EPOCH_REF_TEMP*: 0.00005|\\  
    \verb|LEARNING_RATE: various|\\
    \verb|L2_REG: 1e-5|\\
    \verb|L2_AT_EVAL: False|\\
    \verb|AUGMENT_TRAIN: False|\\
\end{tabular}%
\begin{tabular}[t]{l}
    Find Train/Test Local Minima \\
    \verb|TRAIN_SAMPLE_COUNT: 50000|\\
    \verb|BATCH_SIZE: 10000(res) 5000(vgg)|\\
    \verb|LEARNING_RATE: 1e-4|\\
    \verb|MAX_EPOCH: 10000|\\
    \verb|L2_REG: 1e-5|\\
    \verb|L2_AT_EVAL: True|\\
    \verb|AUGMENT_TRAIN: False|\\
    \verb|TRAIN_SHUFFLE: False|\\
    \verb|TRAIN_DROP_LAST: False|\\
\end{tabular}%

* We scale the number of epochs (\verb|E|) as \verb|E * T = MAX_EPOCHS * MAX_EPOCH_REF_TEMP|.
\end{small}


\subsubsection*{Figure \ref{fig:2basin_g_A} and \ref{append:more_synth_exp}.\ref{fig:1_3_basin}}
The \emph{objective} plots show the synthetic train and test objective (loss) functions $U^\textit{tr}(\theta)$ and $ U(\theta)$. For the \emph{steady-state distribution} plots, we plug the train loss $U^\textit{tr}(\theta)$, the gradient variance $D^\textit{tr}(\theta)$, and the current temperature into (\ref{eq:sgdss}), and evaluate as a function of $\theta$. For the \emph{probability of each basin} plots, we integrate the steady-state distribution over the basin of each minimum (i.e.\ between the two local maxima adjacent to the minimum, or the range endpoint -- noting that $\rho \to 0$ at the range endpoints); the probabilities are plotted as a function of temperature. Finally, the \emph{train vs. test loss} plots show the train and test losses as a function of temperature, i.e.\ train loss is computed as $\int \rho(\theta) U^\textit{tr}(\theta) d\theta)$ (where $\rho(\theta)$ depends on the temperature per (\ref{eq:sgdss}).

\begin{small}
\begin{tabular}{lllll}
Param & Figure \ref{fig:2basin_g_A} Left & Figure \ref{fig:2basin_g_A} Right & Figure \ref{append:more_synth_exp}.\ref{fig:1_3_basin} Left & Figure \ref{append:more_synth_exp}.\ref{fig:1_3_basin} Right\\
\verb|seed|   & \verb|0| & \verb|0| & \verb|0| & \verb|0| \\
\verb|minima| & \verb|[-1  1]| & \verb|[-1  1]| & \verb|[0]| & \verb|[-1 0 1]| \\
\verb|weights| & \verb|[0.021 0.1]| & \verb|[0.019 0.1]| & \verb|[0.1]| & \verb|[0.1 0.051 0.3]| \\
\verb|sigmas| & \verb|[0.1 0.5]| & \verb|[0.1 0.5]| & \verb|[0.1]| & \verb|[0.1 0.05 0.3]| \\
\verb|c|      & \verb|0.001| & \verb|0.001| & \verb|0.001| & \verb|0.001| \\
\verb|stddev_shift| & \verb|0.1| & \verb|0.1| & \verb|0.03| & \verb|0.1| \\
\verb|lscale| & \verb|1.0| & \verb|1.0| & \verb|1.0| & \verb|1.0| \\
\verb|wscale| & \verb|0.0| & \verb|0.0| & \verb|0.0| & \verb|0.0| \\
\end{tabular}
\end{small}

\subsubsection*{Figure \ref{fig:vgg_lb_init_hpuh5hk7ja}}
We used a slightly modified version of the code accompanying \cite{li2017visualizing} for training a VGG9 network on CIFAR10. We changed the learning rate schedule to drop at 0.5, 0.75, and 0.9 of total epochs (rather than at fixed epochs), and also added support for reshuffling the train and test sets, and using a subsample of the full dataset.
\begin{small}
\begin{tabular}[t]{l}
  Stage 1 hyperparameters \\
  \verb|rand_seed: range(0, 10)|\\
  \verb|batch_size: 4096|\\
  \verb|lr: 0.1|\\
  \verb|lr_decay: 0.1|\\
  \verb|epochs: 300|\\
  \verb|weight_decay: 0.0005|\\
  \verb|momentum: 0.9|\\
\end{tabular}%
\begin{tabular}[t]{l}
  Stage 2 hyperparameters:\\
  \verb|resume_epoch: 1200|\\
  \verb|rand_seed: -1|\\
  \verb|batch_size: various|\\
  \verb|lr: various|\\
  \verb|lr_decay: 1.0|\\
  \verb|epochs: 2200|\\
  \verb|weight_decay: 0.0005|\\
  \verb|momentum: 0|\\
\end{tabular}
\begin{tabular}[t]{l}
  Figure \ref{fig:vgg_lb_init_hpuh5hk7ja} right \\
  \verb|rand_seed: -1|\\
  \verb|batch_size: 3000|\\
  \verb|lr: 0.1|\\
  \verb|lr_decay: 0.1|\\
  \verb|epochs: 2000|\\
  \verb|weight_decay: 0.0005|\\
  \verb|momentum: 0|\\
  \verb|shuffle_seed: 0:25)|\\
  \verb|sample_size: 10000:50000|\\
\end{tabular}
\end{small}

\end{document}